\documentclass[11pt]{article}

\usepackage[preprint]{acl}

\usepackage{times}
\usepackage{latexsym}
\usepackage{enumitem}
\usepackage{amsmath}
\usepackage{amssymb}
\usepackage{graphicx}
\usepackage{subcaption}
\usepackage[capitalize]{cleveref}
\usepackage{booktabs}

\usepackage[T1]{fontenc}

\usepackage[utf8]{inputenc}

\usepackage{microtype}

\usepackage{inconsolata}

\title{A Graph Signal Processing Perspective on Numerical Sequence Representations in LLM In-Context Learning}

\author{
  \textbf{Jiajun Bao\textsuperscript{1}\NoHyper\thanks{Correspondence: \texttt{jb2777@cornell.edu}.\newline Our code and data are publicly available at \url{https://github.com/Jiajun-Bao/LLM-Graph-Signal-Processing}.}\endNoHyper,\enspace
  Zihao Qi\textsuperscript{1},\enspace
  Toni J.B. Liu\textsuperscript{1},\enspace
  Gurbir Arora\textsuperscript{1}}\\[0.15em]
  \textbf{Rapha\"el Sarfati\textsuperscript{2},\enspace
  Nicolas Boull\'e\textsuperscript{3},\enspace
  Christopher J. Earls\textsuperscript{1,2}}\\[0.25em]
  \textsuperscript{1}Cornell University \quad
  \textsuperscript{2}Goodfire AI \quad
  \textsuperscript{3}Imperial College London
}

\begin{document}
\maketitle
\begin{abstract}
Pretrained large language models (LLMs) have demonstrated in-context learning (ICL) capabilities for numerical inference over sequences serialized as text. Prior work has identified and characterized this form of numerical inference primarily through output-level evaluations such as prediction error. However, how numerical information is organized within LLM representations remains much less understood. To study this internal organization, we adopt a graph signal processing perspective in which attention induces a weighted graph over tokens, while token hidden states define signals on its nodes. Quantitative graph-spectral diagnostics and qualitative token-graph visualizations reveal that representations become more clearly differentiated by input dynamical complexity as context length increases. Simpler inputs produce attention-induced token graphs with stronger global connectivity and smoother, spectrally concentrated hidden-state signals, whereas more complex inputs produce more localized graphs and hidden-state signals with broader spectral support and greater high-frequency energy. These diagnostics and visualizations offer an interpretable view of the internal structure associated with numerical ICL, beyond what output-level prediction error alone reveals. Together, these findings point to systematic, context-dependent internal signatures associated with numerical ICL that are qualitatively conserved across model families.
\end{abstract}

\begin{figure*}[t]
  \centering
  \includegraphics[width=\textwidth]{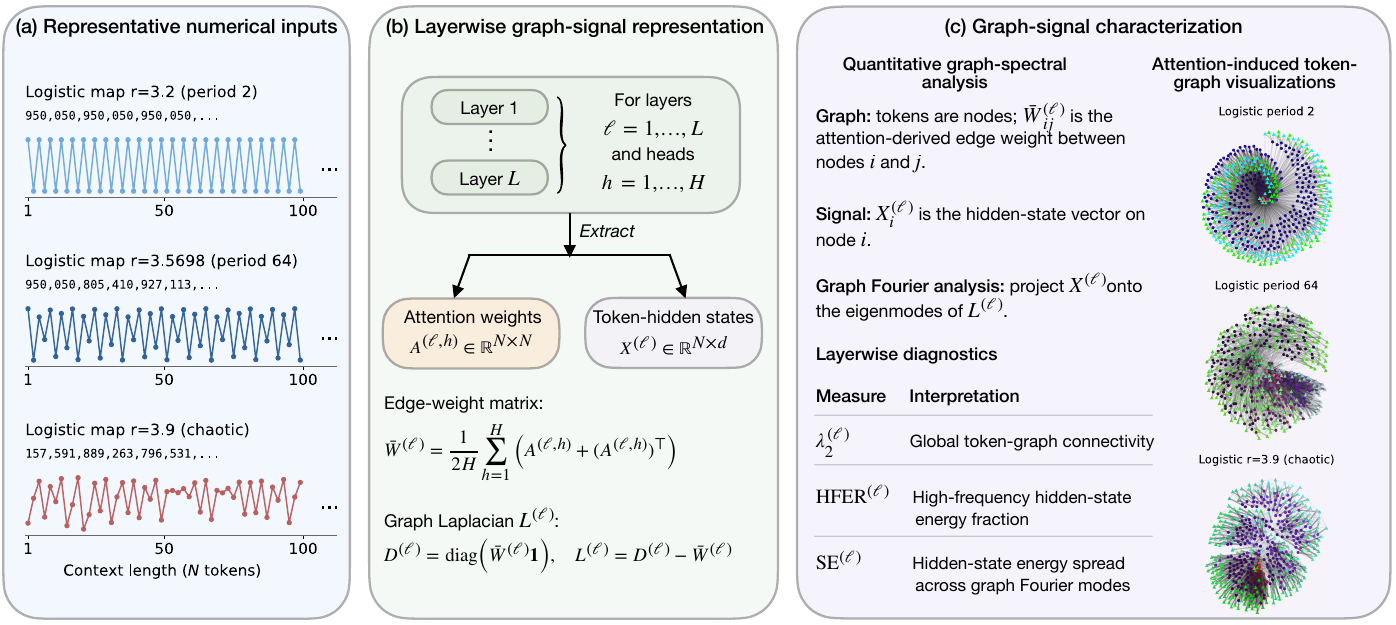}
  \caption{Overview of the graph signal processing framework for studying LLM internal representations during numerical in-context learning. \textbf{(a)} Numerical trajectories are quantized and serialized as comma-delimited sequences of three-digit numbers. \textbf{(b)} At each layer $\ell$, head-aggregated attention defines a weighted, undirected token graph $\bar W^{(\ell)}$ over the $N$ token nodes, hidden states $X^{(\ell)}$ define $d$-dimensional node signals, and Laplacian eigenvectors form the graph Fourier basis, enabling spectral analysis of these signals. \textbf{(c)} Layerwise graph-spectral diagnostics and attention-induced token-graph visualizations characterize these representations. Node positions reflect attention topology, node colors encode hidden-state signals, and circles and triangles denote numeric and separator tokens, respectively. The examples are from the final layer of Llama-3.2-3B at $N=500$.}
  \label{fig:comprehensive-overview}
\end{figure*}

\section{Introduction}
\label{sec:introduction}
Large language models (LLMs) are increasingly studied for numerical inference over sequences serialized as text. Without task-specific parameter updates, they can infer and extrapolate numerical processes, a capability we refer to as numerical in-context learning (ICL). Prior work has demonstrated this capability in time-series forecasting \citep{gruver2024largelanguagemodelszeroshot,jin2024timellm}, 
numerical regression \citep{vacareanu2024fromwordstonumbers,requeima2024llmprocessesnumericalpredictive}, dynamical-system inference \citep{Liu_2024,zhang2025zeroshot}, density estimation \citep{liu2025density}, Markov-model prediction \citep{dai2025pretrainedlargelanguagemodels,zekri2025largelanguagemodelsmarkov}, and partial differential equation extrapolation \citep{bao2026texttrained}. Depending on the task, longer contexts often yield more accurate point predictions, predictive distributions closer to the ground-truth distribution, or more faithful recovery of latent dynamics. Yet most existing studies characterize these improvements through model outputs, treating the model as a black box and leaving it unclear how its internal representations change as context length increases.

\paragraph{} Recent studies have begun to examine internal representations associated with numerical ICL, providing evidence that LLM hidden states systematically encode numerical information. In particular, \citet{sarfati2026shape} identify curved manifolds encoding beliefs about Gaussian parameters, while \citet{piskorz2026eliciting} use trained probes to decode point predictions and uncertainty from LLM hidden states.
Relatedly, \citet{kadlcik2025pretrained} apply trained sinusoid-aware probes to show that numerical values are decodable from input token embeddings, and \citet{stefanik2026language} extend this finding to contextual hidden states across layers.
Motivated by these findings, we jointly analyze token-level hidden representations and the structure of attention-induced token graphs. Specifically, we ask how their organization varies with context length and input dynamical complexity.

\paragraph{} To study this interplay, we adopt the graph signal processing perspective summarized in \cref{fig:comprehensive-overview}: at each transformer layer, attention induces a weighted graph over tokens, while the corresponding token hidden states define signals on its nodes. Our framework supports two complementary analyses: 1) graph-spectral diagnostics, which quantify global attention-graph connectivity, high-frequency energy in hidden-state signals, and the breadth of their spectral support; 2) attention-induced token-graph visualizations, which provide a qualitative view of token-level organization consistent with the quantitative trends.
Across both analyses, we find that, as context length increases, the model's internal representations become more clearly ordered by input dynamical complexity. The diagnostics further distinguish input families whose output-level extrapolation errors remain tightly clustered, suggesting that representation-level analysis can reveal structure not apparent from extrapolation error alone (cf. \cref{fig:extrapolation-scaling-error} in \cref{app:numerical-inputs-extrapolation}).

\section{Background and Related Work}
\paragraph{Graph spectra and graph signals.}
Spectral graph theory studies graph structure through operators such as the Laplacian, whose spectrum captures connectivity, cuts, and global organization \citep{fiedler1973algebraic,chung1997spectral}. Graph signal processing extends this spectral viewpoint to data defined on graph nodes: for undirected graphs, Laplacian eigenvectors form a graph Fourier basis whose modes are ordered by how rapidly nodewise signals vary over the graph \citep{hammond2011wavelets,shuman2013emerging,ortega2018graph}. This classical framework relates the variation of graph-supported signals to the geometry of the underlying graph.

\paragraph{Graph-spectral analysis of LLMs.}
Recent graph-spectral work on LLMs has focused on hallucination detection and reasoning verification. \citet{binkowski2025hallucination} use attention-derived Laplacian spectra for hallucination detection. Closer to our setting, \citet{noel2025graph,noel2026geometry} treats attention-induced token graphs together with token-level hidden-state signals, using diagnostics such as spectral entropy, high-frequency energy ratio, and the Fiedler value for hallucination analysis and proof verification. At the output level, \citet{balaji2026sparse} constructs sentence-level graph signals from generated chain-of-thought traces, enabling verification without access to internal attention weights or hidden states. We adapt closely related diagnostics employed by \citet{noel2026geometry} to numerical ICL and complement them with token-graph visualizations developed in this work to characterize the organization of internal numerical representations.

\paragraph{Applications of graph constructions in LLMs.}
Several recent works represent LLM reasoning traces or token-level attention patterns as graphs, including sentence-level chain-of-thought dependency graphs \citep{bogdan2025thought}, attention-induced token graphs for reinforcement-learning credit assignment \citep{dong2026flowtracer}, and attributed attention graphs used with graph neural networks for hallucination detection \citep{frasca2026neural}. In contrast, we use training-free, layerwise visualizations of attention-induced token graphs to display attention topology alongside low-dimensional hidden-state structure.

\paragraph{Mechanistic interpretability of numerical ICL.}
Alongside the representation-level studies discussed in the introduction, \citet{lan2024towards} use iterative pruning and path patching to identify circuits supporting the continuation of short arithmetic sequences with increasing values. \citet{chowdhury2026mechanistic} use linear probing and activation patching to show that Llama-3.1-8B encodes and causally retrieves first differences in sequences whose increments follow a repeating pattern. Beyond sequence continuation, mechanistic work on pretrained LLMs has studied ICL for three-operand arithmetic \citep{polyakov2025interpretability} and for adding a fixed integer to an input \citep{hu2026understanding}. Complementing these studies, our analysis provides a training-free, label-free characterization of the joint organization of attention topology and hidden-state representations.

\paragraph{} Collectively, these studies leave open how the organization of LLM internal representations during numerical ICL varies with context length and input dynamical complexity. We investigate this question through complementary quantitative and qualitative analyses within a graph signal processing framework.

\section{Methods}
\label{sec:methods}
\subsection{Numerical Inputs and Tokenization}
\label{sec:numerical-input-families-llm-input-preparation}

\paragraph{Controlled numerical input suite.}
Our main experiments use ten one-dimensional sequence families spanning different levels of dynamical complexity under a fixed prompt format: a sampled constant baseline, logistic map trajectories \citep{may1976simple}, and trajectories from the $x$-component of the Lorenz system \citep{lorenz1963deterministic}. The logistic map, $x_{t+1}=r x_t(1-x_t)$, where $x_t\in[0,1]$ and $r \in [0, 4]$ is a control parameter, serves as our primary controlled family. Our selected values of $r$ span periodic regimes with periods $2,4,8,16,32,$ and $64$, followed by two chaotic regimes along the classical period-doubling route \citep{strogatz2024nonlinear}. The logistic map therefore provides a controlled setting for isolating how input dynamics affect LLM representations: varying $r$ produces trajectories with progressively longer periods and then chaotic dynamics while preserving the recurrence form. 

Throughout, we use \emph{dynamical complexity} operationally, with increasing complexity corresponding to the progression from constant inputs through periodic inputs of increasing period to chaotic inputs. This ordering is also supported quantitatively by a standard entropy-based characterization of the input dynamics, as detailed in \cref{app:input-dynamical-complexity}. The constant sequence provides a low-complexity baseline, whereas the Lorenz trajectory provides a complementary observation from a continuous-time chaotic system. We evaluate an ensemble of 20 trajectory realizations per family. \cref{app:numerical-inputs-extrapolation} provides full input-generation details. To broaden the input coverage beyond the controlled dynamical-system suite, we additionally consider structured numerical sequences with distributional shifts, smooth periodic variation, and stochastic evolution, as described in \cref{app:additional-token-graph-visualizations}.

\paragraph{Quantization and serialization.}
Following prior work \citep[e.g.,][]{gruver2024largelanguagemodelszeroshot,Liu_2024}, we encode each numerical input sequence as a quantized, comma-delimited text prompt. For a trajectory $(x_1,\ldots,x_T)$, scalar values are linearly rescaled to the interval $[50,950]$ and rounded to obtain quantized integers $q_i \in \{50,\ldots,950\}$. Each $q_i$ is formatted as a zero-padded three-digit decimal string. The encoded values are separated by commas, yielding strings such as ``\texttt{087,173,642,...}''. To ensure comparability across context lengths, we quantize the full trajectory before truncation so that the numerical mapping remains fixed. We choose this format because widely used tokenizers, such as those used by GPT-4 \citep{openai2023gpt4} and Llama-3-family models \citep{grattafiori2024llama3}, encode three-digit integer strings (\texttt{000}--\texttt{999}) and commas as single tokens. Under this serialization, $T$ numerical states yield $T$ numeric and $T$ separator tokens, so the analyzed context length is $N=2T$. This avoids splitting numerical states across tokens and preserves one-to-one state-to-token alignment for graph analysis. Further quantization and serialization details are provided in \cref{app:numerical-inputs-extrapolation}.

\subsection{Graph-Spectral Diagnostics}
\label{sec:graph-spectral-diagnostics}
We adapt three established spectral diagnostics employed by \citet{noel2026geometry} to numerical ICL: the normalized Fiedler value, high-frequency energy ratio, and spectral entropy (all defined below). In our setting, these diagnostics characterize how global attention-graph connectivity, hidden-state roughness, and spectral diversity vary with input dynamical complexity and context length.

\paragraph{Layerwise representations.}
For an $N$-token input, we record two objects at each layer. First, for each layer $\ell\in\{1,\ldots,L\}$ and attention head $h\in\{1,\ldots,H\}$, let $A^{(\ell,h)}\in\mathbb{R}^{N\times N}$ denote the post-softmax attention matrix, where $A^{(\ell,h)}_{ij}$ is the attention weight from query token $i$ to key token $j$.\footnote{When present, we exclude non-content special tokens introduced during model input preprocessing, such as $\texttt{<BOS>}$, and renormalize each attention row over the remaining tokens.} Second, let $X^{(\ell)}\in\mathbb{R}^{N\times d}$ denote the hidden-state matrix entering layer $\ell$, whose $i$-th row $X_i^{(\ell)}$ is the $d$-dimensional representation of analyzed token $i$. We pair each layer's attention matrices with its input hidden states because those states generate the queries and keys underlying the attention weights.

\paragraph{Graph construction and Fourier modes.}
To support standard Laplacian-based graph-spectral analysis, we construct one undirected weighted graph per layer by representing each analyzed token as a node, symmetrizing each head's attention matrix, and averaging the symmetrized matrices across heads (see Appendix \cref{fig:last-layer-attention-matrices} for representative final-layer attention matrices):
\[
\vspace{-3pt}
\bar W^{(\ell)}
=
\frac{1}{2H}
\sum_{h=1}^{H}
\left(
A^{(\ell,h)}+(A^{(\ell,h)})^\top
\right).
\]
By construction, $\bar W^{(\ell)}_{ij}$ is the attention-derived edge weight between tokens $i$ and $j$. We define the corresponding degree matrix as $D^{(\ell)}=\operatorname{diag}\!\left(\bar W^{(\ell)}\mathbf 1\right)$ and the combinatorial graph Laplacian as $L^{(\ell)}=D^{(\ell)}-\bar W^{(\ell)}$.
The resulting $L^{(\ell)}$ is symmetric positive semidefinite and therefore admits an orthonormal eigendecomposition $L^{(\ell)}=U^{(\ell)}\Lambda^{(\ell)}(U^{(\ell)})^\top$. The eigenvectors in $U^{(\ell)}$ define graph Fourier modes ordered from low to high graph frequency \citep{shuman2013emerging}. Treating the hidden-state matrix $X^{(\ell)}$ as a vector-valued graph signal whose value at node $i$ is $X_i^{(\ell)}$, we project it onto the graph Fourier basis as $\widehat X^{(\ell)}=(U^{(\ell)})^\top X^{(\ell)}$.
Low-eigenvalue modes correspond to smoother hidden-state variation over the attention-induced graph, whereas high-eigenvalue modes correspond to sharper variation across token pairs joined by larger attention weights, as follows from the Laplacian quadratic-form identity in \cref{app:laplacian-quadratic-form}.

\paragraph{Diagnostic 1: Normalized Fiedler value.}
The Fiedler value---the second-smallest eigenvalue of the combinatorial graph Laplacian $L^{(\ell)}$---is a classical measure of global connectivity \citep{fiedler1973algebraic}. To facilitate comparisons across graph sizes and degree scales, we use the standard normalized formulation \citep{chung1997spectral}:
\[
\mathcal L^{(\ell)}
=
(D^{(\ell)})^{-1/2}L^{(\ell)}(D^{(\ell)})^{-1/2}.
\]
The spectrum of $\mathcal L^{(\ell)}$ lies in $[0,2]$. Its second-smallest eigenvalue, denoted by $\lambda_2^{(\ell)}$, is the normalized Fiedler value. Larger $\lambda_2^{(\ell)}$ indicates a more globally integrated attention graph, whereas smaller $\lambda_2^{(\ell)}$ indicates more localized graph structure. \Cref{app:graph-spectral-robustness} examines additional low-end eigenvalues to characterize multiway localized graph structure beyond $\lambda_2^{(\ell)}$ \citep{lee2014multiway}.

\paragraph{Diagnostic 2: High-frequency energy ratio (HFER).}
Let $\widehat{\mathbf x}^{(\ell)}_m\in\mathbb R^d$ denote the graph Fourier coefficient vector for mode $m$, given by the $m$th row of $\widehat X^{(\ell)}$, and $p_m^{(\ell)} = \|\widehat{\mathbf x}^{(\ell)}_m\|_2^2/\sum_{r=1}^{N}\|\widehat{\mathbf x}^{(\ell)}_r\|_2^2$ the normalized modal energy. The HFER is the fraction of hidden-state energy assigned to high graph-frequency modes,
\[
\mathrm{HFER}^{(\ell)}
=\sum_{m=K+1}^{N}p_m^{(\ell)}.
\]
We use $K=\lfloor N/2\rfloor$ as the default cutoff, so HFER measures the fraction of energy in the high-frequency half of the spectrum. \cref{app:graph-spectral-robustness} shows that the qualitative behavior remains consistent across alternative cutoffs. Larger HFER indicates that a greater fraction of hidden-state energy lies in modes associated with sharper variation over the attention-induced graph.

\paragraph{Diagnostic 3: Spectral entropy (SE).}
Whereas HFER measures the fraction of hidden-state energy in high-frequency modes, SE measures how broadly that energy is distributed across graph Fourier modes. Using the normalized modal energies $p_m^{(\ell)}$, we define
\[
\mathrm{SE}^{(\ell)}
=
-\sum_{m=1}^{N}p_m^{(\ell)}\log p_m^{(\ell)}.
\]
We report $\exp(\mathrm{SE}^{(\ell)})$ as the effective spectral support, following the standard effective-number interpretation of exponentiated Shannon entropy \citep{jost2006entropy}. If energy is distributed equally across exactly $k$ modes, then $\exp(\mathrm{SE}^{(\ell)})=k$. More generally, it represents the number of equally weighted graph Fourier modes that would yield the observed spectral entropy. Larger values indicate broader spectral support, whereas smaller values indicate concentration in fewer modes.

\paragraph{Layer averaging.}
In the main text, we report diagnostics averaged across model layers. The corresponding layer-resolved results in \cref{app:layer-resolved-spectral-diagnostics} broadly reproduce the same qualitative trends, supporting layer averaging as a concise and representative summary across model depth. We use $\lambda_2$, $\mathrm{HFER}$, and $\exp(\mathrm{SE})$ to denote the resulting layer-averaged diagnostics.

\subsection{Token-Graph Visualization Pipeline}
\label{sec:graph-visualization-workflow}

\begin{figure*}[t]
  \centering
  \includegraphics[width=\textwidth]{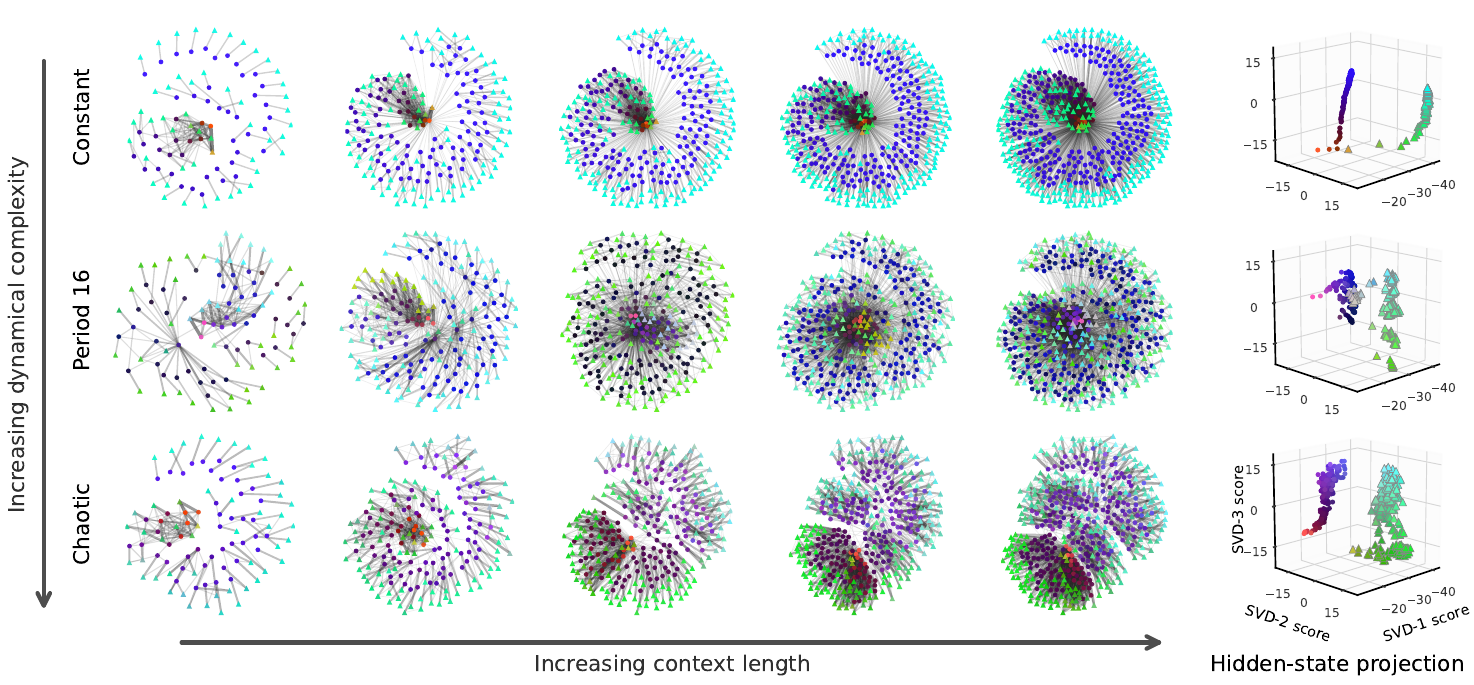}
  \caption{Illustration of attention-induced token graphs across context lengths. Rows show final-layer results from Llama-3.2-3B for representative constant, period-16 logistic map, and chaotic logistic map ($r=4.0$) inputs. The first five columns show token graphs at $N\in\{100,200,300,400,500\}$. Node positions reflect input-specific attention weights through a Fruchterman--Reingold layout, with the same initialization used across inputs at each $N$. Edge thickness encodes attention-derived weight. Only edges above $2.5/N$ are displayed to reduce visual clutter. Node colors encode the leading three uncentered SVD coordinates of token hidden states, with the  $N=500$ projections shown in the rightmost column. Circles mark numeric tokens and triangles mark separator tokens.}
  \label{fig:context-length-graphs-hidden-states}
\end{figure*}

We complement the graph-spectral diagnostics with layerwise two-dimensional graph visualizations in which node color summarizes token hidden-state representations and node position reflects attention topology. These visualizations provide a qualitative view of the organization characterized quantitatively by our spectral diagnostics. \Cref{fig:context-length-graphs-hidden-states} illustrates this joint encoding for three representative input families across context lengths.

\paragraph{Node-color construction for token graphs.}
In classical graph signal processing, a scalar signal is often visualized by mapping its value at each node to color or vertical height \citep{shuman2013emerging}. To obtain a color encoding of the vector-valued token hidden states, we project the hidden-state matrix at each layer onto its leading three uncentered SVD coordinates and assign them to the red, green, and blue (RGB) channels; \cref{sec:attention-graph-construction} provides the full construction. At $N=500$, the leading three components capture 75\% of the hidden-state energy on average across the main numerical input families and analyzed layers (Appendix \cref{fig:cumulative-svd-energy}), supporting their use as a compact representation of the dominant hidden-state structure. Corresponding projections for all ten main numerical input families are provided in Appendix \cref{fig:last-layer-signal-visualization-all-inputs}. For other inputs whose hidden-state energy is distributed across more SVD components, the visualization can be extended with additional color-coded views.

\paragraph{Attention-based node positions.}
Visualizing the attention-induced token graph as a two-dimensional node-link diagram requires choosing a layout. Layouts determined solely by the tokens' sequential positions, such as the spiral initialization shown in Appendix \cref{fig:last-layer-fr-iteration-trajectory}, preserve input order. However, because attention weights do not influence node placement in these layouts, the resulting geometry may obscure nonlocal interactions and community structure reflected in the LLM's internal attention organization. We therefore apply the Fruchterman--Reingold (FR) algorithm \citep{fruchterman1991graph} to the full weighted graph defined by the symmetrized attention matrix $\bar W^{(\ell)}$. The algorithm iteratively updates the node positions by treating nodes as mutually repelling particles, which discourages node overlap, and edges as attractive forces weighted by the corresponding off-diagonal entries of $\bar W^{(\ell)}$, so larger attention-derived edge weights exert stronger attraction. At a fixed $N$, all inputs use the same spiral initialization and layout settings; differences in the relaxed node positions therefore reflect their input-specific attention weights rather than different starting coordinates. Starting from the token-ordered spiral, we run 100 iterations, allowing the layout to relax toward a stable configuration, and use the resulting positions for all subsequent token-graph visualizations. Appendix \cref{fig:last-layer-fr-iteration-trajectory} shows that most visible layout reorganization occurs by 50 iterations, with only minor changes thereafter. \Cref{sec:attention-graph-construction} provides implementation details and demonstrates the robustness of the relaxed layouts to initialization perturbations.

\paragraph{Edge display threshold.}
The FR layout is computed using the full weighted attention graph, which contains $N(N-1)/2$ off-diagonal edges. Rendering every connection would turn the visualization into a dense ``hairball'' in which meaningful patterns are no longer discernible. To reduce visual clutter, we display only edges satisfying $\bar W^{(\ell)}_{ij}>\kappa/N$, using $\kappa=2.5$ for all main visualizations. Appendix \cref{fig:last-layer-graph-threshold-robustness} shows that the qualitative organization is robust to changes in the display threshold. This threshold affects only edge rendering; the FR node positions and all graph-spectral diagnostics are computed from the full weighted graph.

\section{Results}

We examine how the model's internal representations vary with the dynamical complexity of the input across increasing context lengths. \Cref{sec:spectral-diagnostics} quantifies this representational variation using graph-spectral diagnostics, and \Cref{sec:attention-graph-visualizations} visualizes the associated changes in the organization of attention-induced token graphs. 

\paragraph{Choice of model.}
Our main analysis focuses on the pretrained base Llama-3.2-3B model. \cref{app:model-scale-instruction-tuning} extends this analysis to Llama-3.2-1B and Llama-3.1-8B, compares base and instruction-tuned variants, and broadens the evaluation to Microsoft Phi-4 \citep{abdin2024phi4technicalreport} and Hugging Face SmolLM3 \citep{bakouch2025smollm3}. The broad context-length- and complexity-dependent trends persist across these models. Within the evaluated Llama 3 family, the 1B base model distinguishes the dynamical regimes less sharply than the 3B and 8B base models, and the instruction-tuned variants show weaker regime separation than their base counterparts. Within a shared architecture family, these comparisons suggest that model scale and instruction tuning both affect how distinctly dynamical regimes are expressed in the model's internal representations.

\subsection{Graph-Spectral Signatures of Dynamical Complexity Across Context Lengths}
\label{sec:spectral-diagnostics}
For our quantitative graph-spectral analysis, we compute the diagnostics defined in \Cref{sec:graph-spectral-diagnostics} for 20 realizations of each numerical input family introduced in \Cref{sec:numerical-input-families-llm-input-preparation} at 20 context lengths ranging from 100 to 2000 tokens in increments of 100.

\begin{figure*}[t]
  \centering
  \includegraphics[width=\textwidth]{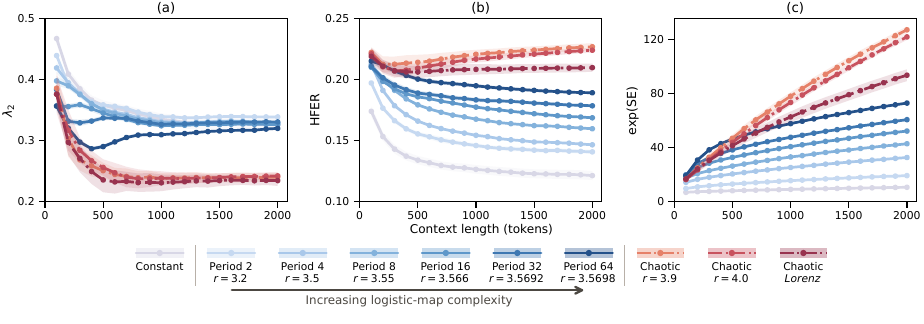}
  \captionsetup{skip=2pt}
  \caption{Layer-averaged graph-spectral diagnostics for the main numerical input families using Llama-3.2-3B. Panels report (a) normalized Fiedler value $\lambda_2$, (b) HFER, and (c) effective spectral support $\exp(\mathrm{SE})$. At longer contexts, chaotic inputs show weaker global attention connectivity (lower $\lambda_2$), greater high-frequency hidden-state energy (higher HFER), and broader effective spectral support (higher $\exp(\mathrm{SE})$) than constant and periodic inputs; periodic families vary progressively with period. Curves and shading show the mean and $\pm 1$ standard deviation across 20 realizations after averaging each diagnostic across transformer layers.}
  \label{fig:layer-averaged-combined-diagnostics}
\end{figure*}

\paragraph{Global attention connectivity reveals context-dependent regime separation.}
\cref{fig:layer-averaged-combined-diagnostics}(a) shows that, with sufficient context, the normalized Fiedler value separates the inputs into two broad groups: the chaotic logistic map and Lorenz inputs exhibit lower $\lambda_2$, whereas the constant and periodic inputs exhibit higher $\lambda_2$. Because lower $\lambda_2$ indicates weaker global connectivity, the chaotic inputs induce more localized attention geometry, while the non-chaotic inputs retain more globally integrated graphs. Consistent with this interpretation, Appendix \cref{fig:layer-averaged-low-end-laplacian-spectrum} shows that the chaotic inputs remain systematically below the non-chaotic inputs across additional low-end eigenvalues, indicating multiway localized graph structure beyond $\lambda_2$. Among the periodic inputs, those with higher periods exhibit a finer dependence on context length. Periods 32 and 64 initially follow the decline of the chaotic inputs before turning upward at $N=300$ and $N=400$, respectively, and approaching the lower-period curves. These ordered turning points suggest that the model needs progressively more context to resolve increasingly complex input dynamics in its internal attention organization.

\paragraph{Graph-signal smoothing varies with dynamical complexity.}
\cref{fig:layer-averaged-combined-diagnostics}(b) shows an increasingly clear complexity-ordered HFER ladder as context length increases. HFER decreases most strongly for the constant and lower-period inputs, the decline weakens as period increases, and the chaotic inputs remain comparatively high after an initial decline. Together, these trends indicate that the effect of additional context on how smoothly hidden states vary over the attention-induced graph depends on input dynamical complexity: for simpler inputs, hidden-state energy becomes increasingly concentrated in low-frequency modes, whereas for more complex inputs, a larger fraction remains in high-frequency modes, preserving sharper variation across the graph.

\paragraph{Effective spectral support expands at complexity-dependent rates.}
\cref{fig:layer-averaged-combined-diagnostics}(c) shows that the effective spectral support, $\exp(\mathrm{SE})$, grows with context length for all inputs but remains well below its theoretical maximum $N$. Thus, increasing context does not simply cause the representation to spread uniformly across the larger set of available spectral modes; instead, it continues to favor a restricted spectral subset. Over the evaluated context range, effective support expands at a complexity-dependent rate: growth is modest for the constant and lower-period inputs, whereas the higher-period and chaotic inputs show stronger expansion. These differences indicate that additional context affects the effective spectral dimensionality of internal representations differently across input families: hidden-state energy for simpler dynamics remains concentrated within a compact, slowly expanding spectral support, whereas for more complex dynamics it is distributed across an increasingly broad set of graph Fourier modes.

\paragraph{}Taken together, the diagnostics reveal an ordered complexity gradient: greater input complexity corresponds to weaker global attention connectivity, higher HFER, and broader effective spectral support, with periodic inputs exhibiting graded changes as their period increases. \cref{app:non-graph-controls} shows that the broad trends also appear in attention-only and hidden-state-only controls, while the graph-spectral diagnostics more clearly separate input families. Additional controls in \cref{app:additional-numerical-controls} show that shuffling numerical states while preserving token identities and frequencies substantially weakens complexity-dependent separation across all three diagnostics, indicating that token diversity alone does not explain the observed organization. The broad context- and complexity-dependent trends persist for numerical-token subgraphs and higher-dimensional inputs. Overall, these results show that the model's internal representations reflect numerical input dynamics in a rich and systematically ordered manner as context length increases.

\subsection{Attention-Induced Token-Graph Organization Across Dynamical Complexity and Context Length}
\label{sec:attention-graph-visualizations}
\begin{figure*}[t]
  \centering
  \begin{minipage}[t]{0.695\textwidth}
    \vspace{0pt}
    \centering
    \includegraphics[width=\linewidth]{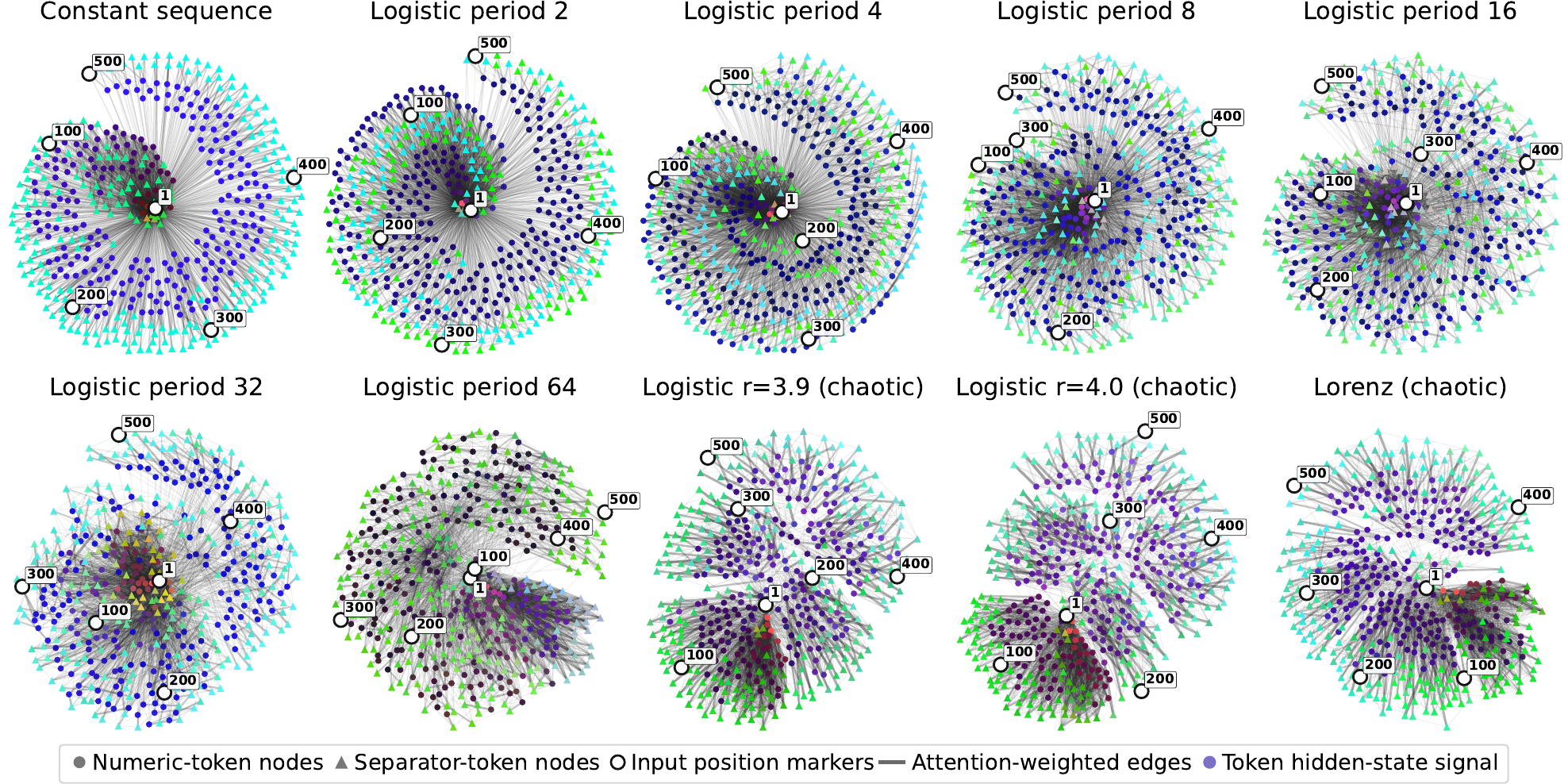}
  \end{minipage}%
  \hspace{0.01\textwidth}%
  \begin{minipage}[t]{0.295\textwidth}
    \vspace{0pt}
    \centering
    \includegraphics[width=\linewidth]{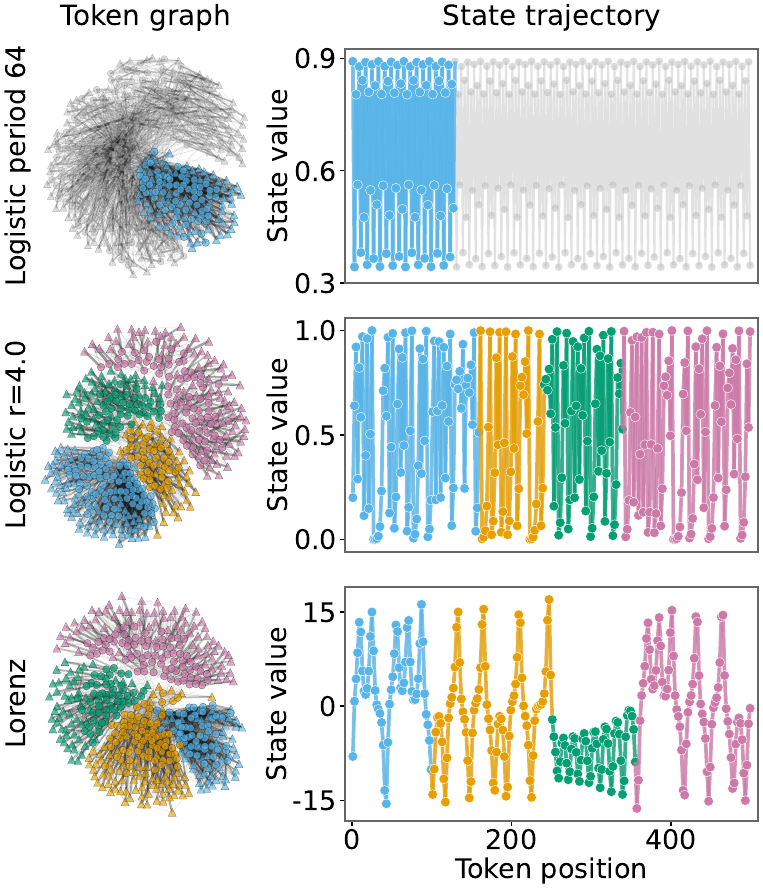}
  \end{minipage}
  \caption{Final-layer attention-induced token graphs and trajectory mappings for Llama-3.2-3B at $N=500$. \textbf{Left}: Node positions reflect the topology of the attention-induced token graph, node colors encode hidden-state graph signals, and circles and triangles denote numeric and separator tokens. Simpler inputs show globally integrated layouts, whereas chaotic inputs show localized, clique-like subnetworks. \textbf{Right}: Matched colors link these clusters to trajectory intervals for period 64, the chaotic logistic map, and Lorenz inputs. For period 64, a clique-like cluster appears early, before sufficient context reveals the recurrence, but no comparable clusters appear later; chaotic inputs exhibit multiple such clusters throughout.}
  \label{fig:last-layer-graph}
  \vspace{-1pt}
\end{figure*}

Following the construction in \Cref{sec:graph-visualization-workflow}, \cref{fig:last-layer-graph} presents representative final-layer attention-induced token graphs at $N=500$. At this context length, the layer-averaged $\lambda_2$ trends in \cref{fig:layer-averaged-combined-diagnostics}(a) already show broad separation between chaotic and non-chaotic inputs in global attention-graph connectivity. \Cref{fig:context-length-graphs-hidden-states} further shows that the corresponding complexity-dependent visual organization is visible by $N=300$ and remains qualitatively stable thereafter. We focus on the final layer because its attention-defined topology is closest to the representations used to compute the output logits. \Cref{sec:token_graphs} provides complementary visualizations of additional transformer layers.

\paragraph{Token graphs become more localized with dynamical complexity.}
The layouts progress from comparatively regular, globally integrated organization for constant and lower-period inputs to less homogeneous organization as the period increases. The chaotic logistic map and Lorenz inputs contain multiple localized, clique-like subnetworks, consistent with their lower normalized Fiedler values. Beyond topology, the node colors reveal additional structure in the hidden-state signals: numeric and separator tokens generally occupy distinct regions of the representation. For the period-2 input, separator tokens split into two color groups aligned with the alternating numerical states, suggesting that their hidden states encode the trajectory's alternating phase despite their shared token identity.

\paragraph{Long-period periodicity becomes more distinct from chaos as context increases.}
Period 64 provides an informative intermediate case: despite its periodic dynamics, its graph contains a prominent clique-like subnetwork resembling those of the chaotic inputs. The right panel maps this subnetwork and those in the chaotic logistic map and Lorenz inputs to their corresponding trajectory intervals. For period 64, this subnetwork is concentrated within roughly the first $128$ tokens, before the model has observed enough recurrence to distinguish its periodic structure from chaotic dynamics; later trajectory segments do not form similarly pronounced subnetworks. In contrast, the chaotic logistic map and Lorenz graphs contain multiple localized subnetworks throughout the context, each associated primarily with a contiguous trajectory interval. This difference suggests that additional context resolves the early ambiguity of the long-period input, whereas temporal partitioning persists for chaotic dynamics.

\paragraph{} The visualizations complement the spectral diagnostics by showing a broad shift from globally integrated to more localized token-graph organization as dynamical complexity increases. Appendix~\ref{app:additional-token-graph-visualizations} visualizes high-frequency hidden-state components on token graphs, qualitatively complementing the HFER trends, and examines additional structured numerical inputs beyond the controlled dynamical-system suite, including distributional shifts accompanied by distinct token-graph clustering.

\section{Conclusion}
\label{sec:conclusion}

We adopted a graph signal processing perspective to study numerical ICL, in which attention defines a token graph and hidden states form signals on its nodes. As context grows, our graph-spectral diagnostics and analyses of attention-induced token graphs reveal increasingly clear differences across trajectories of varying dynamical complexity. These findings show that the models' emergent capacity for numerical ICL manifests not only in the autoregressive continuation of numerical sequences at the output level but also in the fine-grained, context-dependent organization of their internal representations. More broadly, the framework provides a mesoscale characterization of LLM attention topology and hidden-state organization, complementing microscale neuron- and circuit-level analyses and macroscale output-based evaluations.

\paragraph{} Our study focuses on standard Laplacian-based graph signal processing, using undirected token graphs constructed from symmetrized attention matrices. The resulting Laplacian provides a natural measure of signal smoothness and an orthonormal eigenbasis for graph Fourier analysis. Future work could draw on directed spectral graph theory \citep{chung2005laplacians} and directed graph signal processing \citep{marques2020signal} to examine attention directionality and reveal complementary aspects of internal organization. Another direction is to investigate head-level structure within the existing framework, potentially offering additional insight into the roles and coordination of individual attention heads and how their attention characteristics relate to hidden-state organization.

\paragraph{} In numerical ICL, an important open question is whether the observed spectral signatures and attention-graph locality causally contribute to sequence continuation. Appendix~\ref{app:layer-skipping} provides initial interventional evidence relating graph-spectral diagnostics to next-state extrapolation behavior. Reconstructions from individual graph Fourier modes could localize hidden-state variation, while frequency-band ablations could test the functional importance of particular bands. Contrasting locality-preserving with locality-disrupting attention-edge interventions could test the role of graph locality.

\paragraph{} Beyond numerical sequences, preliminary analyses in \cref{sec:additional-language-inputs} show that, at matched context lengths, natural-language and code inputs consistently exhibit more localized attention topology, a greater high-frequency share of hidden-state energy, and broader effective spectral support than numerical inputs. Across natural languages, programming languages, and English-language domains, it remains to determine how input properties such as syntax, semantic content, formatting, and predictability, and model-dependent factors such as tokenization granularity and training-data exposure contribute to both the observed separation from numerical inputs and graph-spectral variation within these broader input classes.

\section*{Limitations}
\paragraph{Mechanistic scope.}
Our graph-spectral diagnostics of attention-induced token graphs and hidden-state signals identify structural and representational signatures associated with numerical ICL, but do not establish the causal mechanisms that produce this behavior.

\paragraph{Model and tokenization choices.}
Our experiments cover several model families with Llama 3--style tokenization, for which each three-digit state corresponds to a single numeric node in the attention-induced token graph and each comma delimiter to a separator node. Other model families use different numerical tokenization schemes; for example, Gemma 4 tokenizes each digit separately \citep{gemmateam2026gemma4}. Applying the same state-aligned construction to such models would require aggregating digit-level tokens into state-level nodes, which we leave to future work.

\paragraph{Access to internal representations.}
The proposed diagnostics and visualizations require access to layerwise attention matrices and hidden-state activations for all analyzed input tokens. Hosted interfaces to frontier models, such as chat applications and APIs, typically do not expose these quantities. The framework is currently most directly applicable to open-weight models whose internal representations are accessible.

\section*{Acknowledgments}
The authors are grateful to Valentin No\"el and David Bindel for helpful discussions, and to the anonymous reviewers for their constructive feedback. This research is funded in part by the Gordon and Betty Moore Foundation through Grant GBMF13901 to Cornell to support the work of G.A.

The authors used OpenAI Codex to assist with code development, improve the paper's readability, and correct typos and grammatical errors. The authors assume responsibility for all content.


\bibliography{custom}

\appendix

\section*{Appendix Contents}
{
\raggedright
\begin{description}[
    leftmargin=1.8em,
    labelwidth=1.2em,
    labelsep=0.6em,
    itemsep=0.15em,
    topsep=0.3em,
    font=\normalfont\bfseries
]
    \item[\ref*{app:numerical-inputs-extrapolation}.]
    \hyperref[app:numerical-inputs-extrapolation]
    {Main Numerical Input Suite and Next-State Extrapolation}

    \item[\ref*{app:input-dynamical-complexity}.]
    \hyperref[app:input-dynamical-complexity]
    {Entropy-Based Quantification of Input Dynamical Complexity}

    \item[\ref*{app:laplacian-quadratic-form}.]
    \hyperref[app:laplacian-quadratic-form]
    {Laplacian Quadratic Forms and Graph Fourier Modes}

    \item[\ref*{app:graph-spectral-robustness-layerwise-consistency}.]
    \hyperref[app:graph-spectral-robustness-layerwise-consistency]
    {Robustness and Layerwise Consistency of Graph-Spectral Diagnostics}

    \item[\ref*{sec:attention-graph-construction}.]
    \hyperref[sec:attention-graph-construction]
    {Attention-Induced Token Graphs: Construction and Visualization Robustness}

    \item[\ref*{sec:token_graphs}.]
    \hyperref[sec:token_graphs]
    {Attention-Induced Token-Graph Visualizations at Additional Transformer Layers}

    \item[\ref*{app:model-scale-instruction-tuning}.]
    \hyperref[app:model-scale-instruction-tuning]
    {Graph-Spectral Diagnostics Across Model Families, Scales, and Instruction Tuning}

    \item[\ref*{app:non-graph-controls}.]
    \hyperref[app:non-graph-controls]
    {Attention-Only and Hidden-State-Only Non-Graph Baselines}

    \item[\ref*{app:additional-token-graph-visualizations}.]
    \hyperref[app:additional-token-graph-visualizations]
    {Additional Attention-Induced Token Graph Visualizations}

    \item[\ref*{app:additional-numerical-controls}.]
    \hyperref[app:additional-numerical-controls]
    {Additional Controls and Higher-Dimensional Inputs}

    \item[\ref*{app:layer-skipping}.]
    \hyperref[app:layer-skipping]
    {Preliminary Layerwise Intervention Analysis}

    \item[\ref*{sec:additional-language-inputs}.]
    \hyperref[sec:additional-language-inputs]
    {Preliminary Graph-Spectral Comparison of Numerical, Natural-Language, and Code Inputs}
\end{description}
}

\crefname{appendix}{Appendix}{Appendices}
\Crefname{appendix}{Appendix}{Appendices}

\section{Main Numerical Input Suite and Next-State Extrapolation}
\label[appendix]{app:numerical-inputs-extrapolation}

\paragraph{Main numerical input suite.}
The main numerical input suite comprises ten one-dimensional trajectory families: a constant-sequence baseline whose value is sampled separately for each realization, six periodic logistic map families with periods $2,4,8,16,32,$ and $64$, two chaotic logistic map families, and a family of trajectories obtained from the Lorenz $x$-component. For each family, we construct an ensemble of 20 trajectory realizations. The same realizations are used in the graph-spectral analysis presented in the main text and in the extrapolation analysis described below. Realization $j$ uses a shared initial-state draw $x_0^{(j)}\sim\mathrm{Unif}(0.3,0.7)$ across these families.
For the constant baseline, realization $j$ is assigned the quantized state
\[
q_{\mathrm{const}}^{(j)}
=
\operatorname{round}\!\left(50+900x_0^{(j)}\right).
\]
This quantized value is repeated across all 1000 states of realization $j$ and varies across realizations. Logistic map inputs are generated from $x_{t+1}=r x_t(1-x_t)$. The periodic cases use $r=3.2$, $3.5$, $3.55$, $3.566$, $3.5692$, and $3.5698$, corresponding to periods $2,4,8,16,32,$ and $64$, and the chaotic cases use $r=3.9$ and $r=4.0$. For each realization of a periodic case with expected period $p$, we generate a 20,000-state trajectory, identify the earliest point after which $|x_{t+p}-x_t|<10^{-10}$ holds for the remainder of the trajectory, and extract a 1000-state window after applying a random phase offset in $\{0,\ldots,p-1\}$. We use 20 realizations primarily to characterize variation in the graph-spectral diagnostics for families with greater across-realization trajectory diversity, particularly the higher-period and chaotic families. We retain the same ensemble size for the lower-period families to ensure consistent aggregation across inputs, although a converged period-$p$ orbit has only $p$ possible phases, so duplicate phase selections are expected. For the chaotic logistic families, we instead use direct 1000-state trajectories from the realization-specific $x_0^{(j)}$.
The Lorenz trajectory family is generated from
\[
\begin{array}{rcl}
\dot{x} &=& \sigma(y-x),\\
\dot{y} &=& x(\rho-z)-y,\\
\dot{z} &=& xy-\beta z,
\end{array}
\]
with $\sigma=10$, $\rho=28$, and $\beta=8/3$, the classical chaotic parameter setting \citep{lorenz1963deterministic}. For realization $j$, the initial condition is $(x,y,z)=(-8.0+(x_0^{(j)}-0.5),\,7.0,\,27.0)$, varying the initial $x$-coordinate around the standard $(-8,7,27)$ initialization \citep{brunton2016discovering}. We integrate over $\tau\in[0,100]$ using SciPy's \texttt{solve\_ivp} with its default solver settings, record 1000 evenly spaced time points, and retain only the $x$-component, yielding, for each realization, a partially observed one-dimensional trajectory from a continuous-time chaotic system.

\paragraph{Trajectory quantization and prompt serialization.}
For a logistic or Lorenz realization $\mathbf{u}=(u_1,\ldots,u_{1000})$, let
\[
u_{\min}=\min_i u_i
\qquad\text{and}\qquad
u_{\max}=\max_i u_i.
\]
If $u_{\max}=u_{\min}$, we set $q_i=500$. Otherwise, the quantized state at position $i$ is
\[
q_i
=
\operatorname{round}\!\left(
50+900\frac{u_i-u_{\min}}{u_{\max}-u_{\min}}
\right).
\]
Here, $\operatorname{round}(\cdot)$ denotes rounding to the nearest integer, so $q_i\in\{50,\ldots,950\}$. The sampled constant baseline bypasses this mapping and uses $q_{\mathrm{const}}^{(j)}$ as defined above. The scaling bounds are computed once from the full 1000-state realization before context truncation. Each quantized state is formatted as a zero-padded three-digit string and followed by a comma delimiter, including the final state. Thus, a realization is serialized schematically as ``$q_1,q_2,\ldots,q_{1000},$''.

\begin{figure}
  \centering
  \includegraphics[width=\columnwidth]{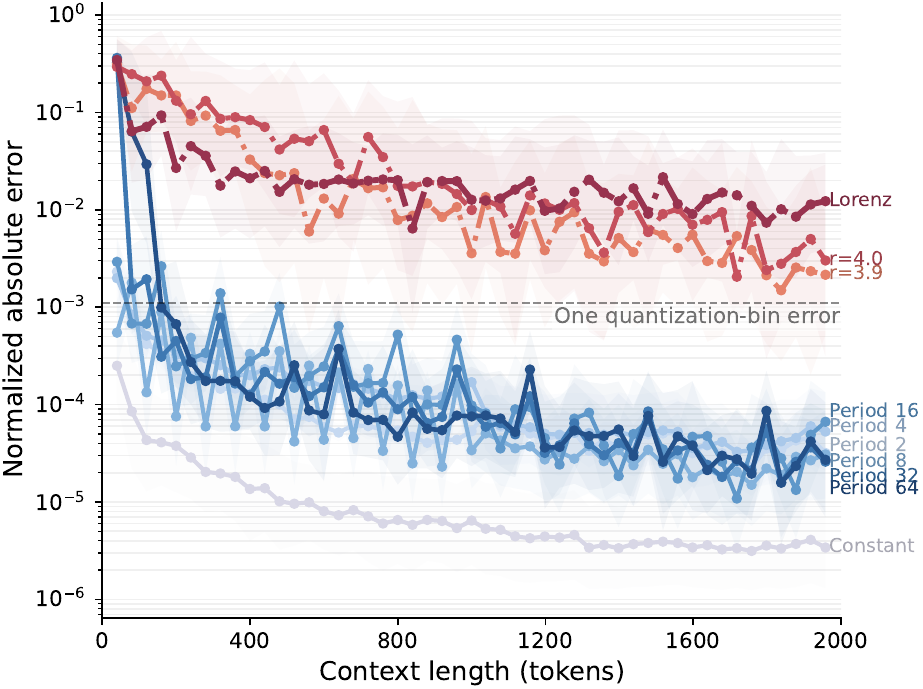}
  \caption{Context-length scaling of next-state extrapolation error with Llama-3.2-3B. Curves show normalized expected-value error for predicting the next quantized state from the preceding serialized numerical context. Longer context generally reduces extrapolation error across the main numerical input families. Several low-error inputs have tightly clustered output-level curves, motivating the representation-level analysis: graph-spectral diagnostics reveal more ordered differences in the model's internal token organization that are less apparent from extrapolation error alone. The shaded bands show one log standard deviation across realizations.}
  \label{fig:extrapolation-scaling-error}
\end{figure}

\paragraph{Next-state extrapolation evaluation.}
\cref{fig:extrapolation-scaling-error} reports output-level next-state extrapolation error across context lengths for the main numerical input families. We evaluate 49 context fractions from 2\% through 98\% in increments of two percentage points. For each fraction $c$, the prompt contains $t_c=\operatorname{round}(c(1000-1))$ states, reserving the following state as the prediction target. For input family $f$, realization $j$, and state index $t$, the prompt contains the first $t$ comma-delimited quantized states and the target is the $(t+1)$-st state, denoted $y^f_{j,t+1}\in\{50,\ldots,950\}$. At the final comma position of the prompt, the model produces next-token logits over its full vocabulary. For each integer $v\in\mathcal{V}=\{0,\ldots,999\}$, let $z^f_{j,t}(v)$ denote the logit assigned to the token encoding the zero-padded three-digit representation of $v$. We retain the full numeric output range \texttt{000}--\texttt{999}, so values outside the input interval \texttt{050}--\texttt{950} provide buffers that reduce boundary truncation of probability mass near the quantization boundaries. We then evaluate a softmax distribution restricted to these numeric tokens
\[
p^f_{j,t}(v)=
\frac{\exp z^f_{j,t}(v)}
{\sum_{u\in\mathcal{V}}\exp z^f_{j,t}(u)}.
\]
The corresponding expected next state is
\[
\hat y^f_{j,t+1}=\sum_{v\in\mathcal{V}} v\,p^f_{j,t}(v).
\]
We measure extrapolation error using the normalized absolute difference between the expected next state and the target:
\[
e^f_{j,t}=\frac{|\hat y^f_{j,t+1}-y^f_{j,t+1}|}{950-50}.
\]
The plotted curves are geometric means of $e^f_{j,t}$ over the 20 realizations at each $t$, with shaded bands corresponding to plus or minus one standard deviation on the log-error scale. This behavioral check shows that extrapolation accuracy generally improves with longer context across all main numerical input families, while chaotic logistic and Lorenz inputs exhibit larger extrapolation errors than simpler periodic and constant inputs. At the same time, several low-error inputs have tightly clustered output-level curves, motivating the representation-level analysis in the main text.

The dashed reference line marks one quantization-bin error, $1/900$: because the error is normalized by $950-50=900$, this corresponds to an absolute error of one quantized state unit. Periodic inputs often lie below this line, indicating that the restricted-softmax expected output is within one bin of the ground-truth quantized state; their curves can still decrease with longer context, indicating further reductions in expected-value error even after reaching this output-level scale.

\section{Entropy-Based Quantification of Input Dynamical Complexity}
\label[appendix]{app:input-dynamical-complexity}

An invariant probability measure $\mu$ provides a time-independent summary of state-space occupancy in a dynamical system \citep{walters1982introduction}. Coarse-grained entropies of invariant state distributions are a classical tool for characterizing the statistical structure of dynamical systems \citep{farmer1982information}. We use Shannon entropy \citep{shannon1948mathematical} to summarize how broadly this distribution is spread across the visited states. This quantity provides quantitative support for the operational ordering of dynamical complexity used in \cref{sec:numerical-input-families-llm-input-preparation}.

For a period-$p$ orbit with distinct states $x_1,\ldots,x_p$, the invariant measure assigns probability $1/p$ to each state:
\[
\mu_p=\frac{1}{p}\sum_{j=1}^{p}\delta_{x_j},
\]
where $\delta_{x_j}$ denotes a unit point mass at $x_j$. Its state-distribution entropy is therefore
\[
H_{\mathrm{state}}(\mu_p)
=
-\sum_{j=1}^{p}\frac{1}{p}\log_2\frac{1}{p}
=
\log_2 p.
\]
Thus, the constant input has entropy $0$, while periods $2,4,8,16,32,$ and $64$ have ideal entropies $1,2,3,4,5,$ and $6$ bits, respectively. To compare all input families on a common basis, we estimate state-distribution entropy directly from each finite quantized trajectory $q_1,\ldots,q_T$ used as an LLM input, where $q_t\in\{50,\ldots,950\}$, using 100 equal-width bins with shared edges spanning $[49.5,950.5]$ across all input families and realizations. Letting $B_b$ denote the $b$-th bin, we compute
\[
\begin{aligned}
\widehat p_{T,b}
&=
\frac{1}{T}\sum_{t=1}^{T}\mathbf{1}\{q_t\in B_b\},\\
\widehat H_T
&=
-\sum_{b:\widehat p_{T,b}>0}
\widehat p_{T,b}\log_2\widehat p_{T,b}.
\end{aligned}
\]

\begin{table}
  \centering
  \small
  \begin{tabular*}{\columnwidth}{@{\extracolsep{\fill}}lr@{}}
    \toprule
    \textbf{Input type} & \textbf{Mean $\boldsymbol{\pm}$ SD (bits)} \\
    \midrule
    Constant sampled                         & $0.000 \pm 0.000$ \\
    Logistic $r=3.2$ (period 2)              & $1.000 \pm 0.000$ \\
    Logistic $r=3.5$ (period 4)              & $2.000 \pm 0.000$ \\
    Logistic $r=3.55$ (period 8)             & $3.000 \pm 0.000$ \\
    Logistic $r=3.566$ (period 16)           & $3.625 \pm 0.000$ \\
    Logistic $r=3.5692$ (period 32)          & $4.039 \pm 0.001$ \\
    Logistic $r=3.5698$ (period 64)          & $4.293 \pm 0.002$ \\
    Logistic $r=3.9$ (chaotic)               & $6.222 \pm 0.057$ \\
    Logistic $r=4.0$ (chaotic)               & $6.295 \pm 0.053$ \\
    Lorenz                                   & $6.360 \pm 0.022$ \\
    \bottomrule
  \end{tabular*}
  \caption{Coarse-grained state-distribution entropy of the quantized numerical input families introduced in \cref{sec:numerical-input-families-llm-input-preparation}. Values are reported as the mean $\pm$ sample standard deviation in bits across 20 realizations at trajectory length $T=1000$, estimated using 100 equal-width bins with shared edges across all input families and realizations.}
  \label{tab:input-state-distribution-entropy}
\end{table}

Applying this estimator to realization $j$ of input family $f$ gives $\widehat H^f_{j,T}$. \Cref{tab:input-state-distribution-entropy} reports the mean and sample standard deviation of $\widehat H^f_{j,1000}$ over $j=1,\ldots,20$ for each family $f$. The empirical entropies are consistent with the intended complexity ordering: entropy increases with logistic map period, while the chaotic logistic and Lorenz inputs exhibit the highest state-distribution entropies. The higher-period values fall below the ideal values $\log_2 p$ because the orbit states are nonuniformly spaced across state space, causing multiple distinct states to fall within the same histogram bin. The reported quantity is therefore a coarse-grained estimate of the entropy of the invariant state distribution.

\begin{figure}
  \centering
  \includegraphics[width=\columnwidth]{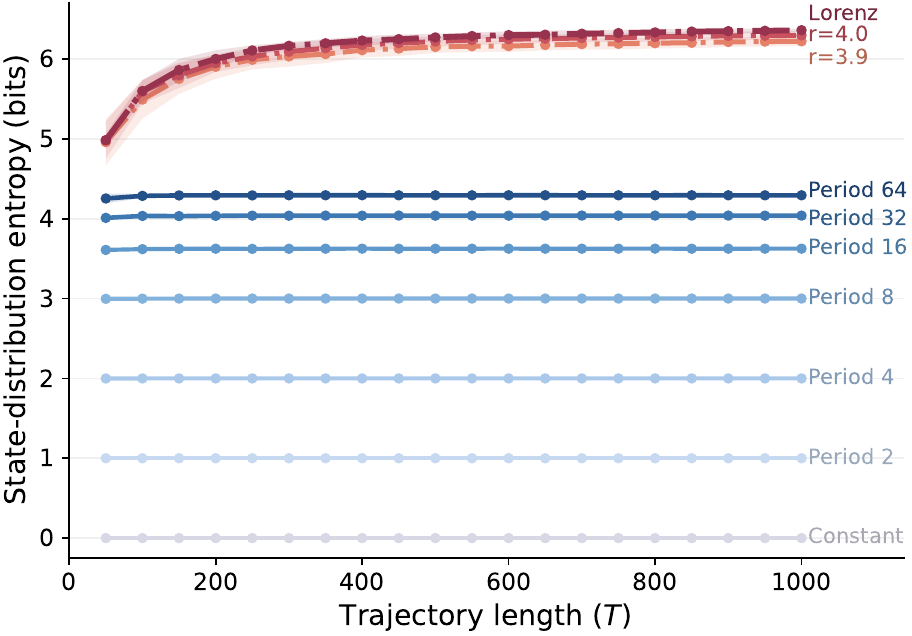}
  \caption{Empirical state-distribution entropy as a function of trajectory length $T$. Periodic estimates stabilize by $T=100$, while chaotic logistic and Lorenz estimates also stabilize relatively early, with little change beyond $T=400$. Entropy is estimated using 100 equal-width bins with shared edges across all input families and realizations. Curves and shading show the mean and $\pm 1$ sample standard deviation across the same 20 realizations used in \cref{sec:numerical-input-families-llm-input-preparation}.}
  \label{fig:input-state-distribution-entropy-context}
\end{figure}

\Cref{fig:input-state-distribution-entropy-context} shows that empirical estimates of state-distribution entropy stabilize with increasing trajectory length, consistent with entropy characterizing the underlying invariant state distribution. The estimates are computed at the context lengths used in the main analysis: $T=50,100,\ldots,1000$ numerical states, corresponding to $N=100,200,\ldots,2000$ tokens under the $N=2T$ serialization described in \cref{sec:numerical-input-families-llm-input-preparation}. The periodic estimates are already close to their $T=1000$ values by $T=100$. The chaotic logistic and Lorenz estimates increase most sharply between $T=50$ and $T=200$, then gradually level off, with only modest changes beyond $T=400$. Thus, the context-dependent growth in the spectral entropy of the LLM's internal representations cannot be explained simply by variation in the estimated state-distribution entropy of the inputs.

\section{Laplacian Quadratic Forms and Graph Fourier Modes}
\label[appendix]{app:laplacian-quadratic-form}

We briefly review why the graph Laplacian measures signal variation over a weighted graph. Let $W\in\mathbb R^{N\times N}$ be a symmetric nonnegative weighted adjacency matrix, let $D=\operatorname{diag}(W\mathbf 1)$, and let $L=D-W$ be the combinatorial graph Laplacian.

First consider a scalar graph signal $\mathbf x=(x_1,\ldots,x_N)^\top$, where $x_i$ is the value assigned to node $i$. The Laplacian quadratic form expands as
\[
\begin{aligned}
\mathbf x^\top L\mathbf x
&=\sum_i D_{ii}x_i^2-\sum_{i,j}W_{ij}x_ix_j\\
&=\frac{1}{2}\sum_{i,j}W_{ij}(x_i-x_j)^2 .
\end{aligned}
\]
Thus, $\mathbf x^\top L\mathbf x$ measures the roughness of the nodewise signal over the graph: differences across node pairs joined by larger weights contribute more strongly to the quadratic form.

This identity also motivates the graph Fourier interpretation of Laplacian eigenvectors. Writing
\[
\begin{aligned}
L&=U\Lambda U^\top,\\
\Lambda&=\operatorname{diag}(\mu_1,\ldots,\mu_N),
\quad
\widehat{\mathbf x}=U^\top\mathbf x,
\end{aligned}
\]
where $0\leq\mu_1\leq\cdots\leq\mu_N$, let $\mathbf u_m$ denote the $m$-th column of $U$. These eigenvectors form the graph Fourier basis used in \Cref{sec:graph-spectral-diagnostics}, and $\widehat{x}_m=\mathbf u_m^\top\mathbf x$ is the graph Fourier coefficient of $\mathbf x$ at mode $m$. The quadratic form can then be written as
\[
\mathbf x^\top L\mathbf x
=\mathbf x^\top U\Lambda U^\top\mathbf x=\widehat{\mathbf x}^\top\Lambda\widehat{\mathbf x}=\sum_{m=1}^{N}\mu_m\widehat{x}_m^2 .
\]
In particular, $\mathbf u_m^\top L\mathbf u_m=\mu_m$, so $\mu_m$ gives the graph variation of the corresponding Fourier mode. Because $L\mathbf 1=\mathbf 0$ and $L$ is positive semidefinite, its smallest eigenvalue is $\mu_1=0$. More generally, the multiplicity of the zero eigenvalue equals the number of connected components of the graph. Thus, for a connected graph, the zero eigenvalue is simple and the second-smallest eigenvalue satisfies $\mu_2>0$. The full weighted attention graphs used in our spectral diagnostics are connected. This connectedness condition applies throughout our analysis. Ordering the eigenvectors by increasing eigenvalue therefore orders the graph Fourier modes from low frequency, corresponding to smooth variation across strongly weighted edges, to high frequency, corresponding to sharper variation.

The same interpretation extends to vector-valued hidden-state signals. Suppose each node $i$ is assigned a hidden-state vector $X_i\in\mathbb R^d$, and collect these vectors as the rows of a matrix $X\in\mathbb R^{N\times d}$. Applying the scalar identity to each hidden dimension gives
\[
\operatorname{tr}(X^\top L X)
=
\frac{1}{2}\sum_{i,j}W_{ij}\|X_i-X_j\|_2^2 .
\]
Equivalently, if $\widehat X=U^\top X$, then
\[
\operatorname{tr}(X^\top L X)
=
\sum_{m=1}^{N}
\mu_m
\|\widehat{\mathbf x}_m\|_2^2 ,
\]
where $\widehat{\mathbf x}_m$ denotes the $m$-th row of $\widehat X$. In our setting, $W=\bar W^{(\ell)}$ is the attention-derived weighted adjacency matrix and $X=X^{(\ell)}$ is the hidden-state matrix entering layer $\ell$. Therefore, the Laplacian quadratic form quantifies how rapidly hidden-state representations vary across attention-weighted token interactions, and the Laplacian eigenvectors provide the graph Fourier modes used in the spectral diagnostics.

Because $U$ is orthonormal, Parseval's identity gives
\[
\|X\|_F^2
=
\|\widehat X\|_F^2
=
\sum_{m=1}^{N}\|\widehat{\mathbf x}_m\|_2^2 .
\]
This identity justifies treating $\|\widehat{\mathbf x}_m\|_2^2$ as the energy assigned to mode $m$ in the HFER and spectral entropy diagnostics.

\section{Robustness and Layerwise Consistency of Graph-Spectral Diagnostics}
\label[appendix]{app:graph-spectral-robustness-layerwise-consistency}
\subsection[appendix]{Robustness of Graph-Spectral Diagnostics}
\label[appendix]{app:graph-spectral-robustness}

\Cref{sec:graph-spectral-diagnostics} defines the normalized Fiedler value as the second-smallest normalized Laplacian eigenvalue and HFER as the fraction of hidden-state energy in the highest-frequency 50\% of graph Fourier modes. Here, we test whether the observed Fiedler-value separation extends to multiway graph structure and whether the HFER trends are robust to the choice of high-frequency cutoff by examining additional low-end eigenvalues and alternative HFER cutoffs.

\begin{figure}
  \centering
  \includegraphics[width=\columnwidth]{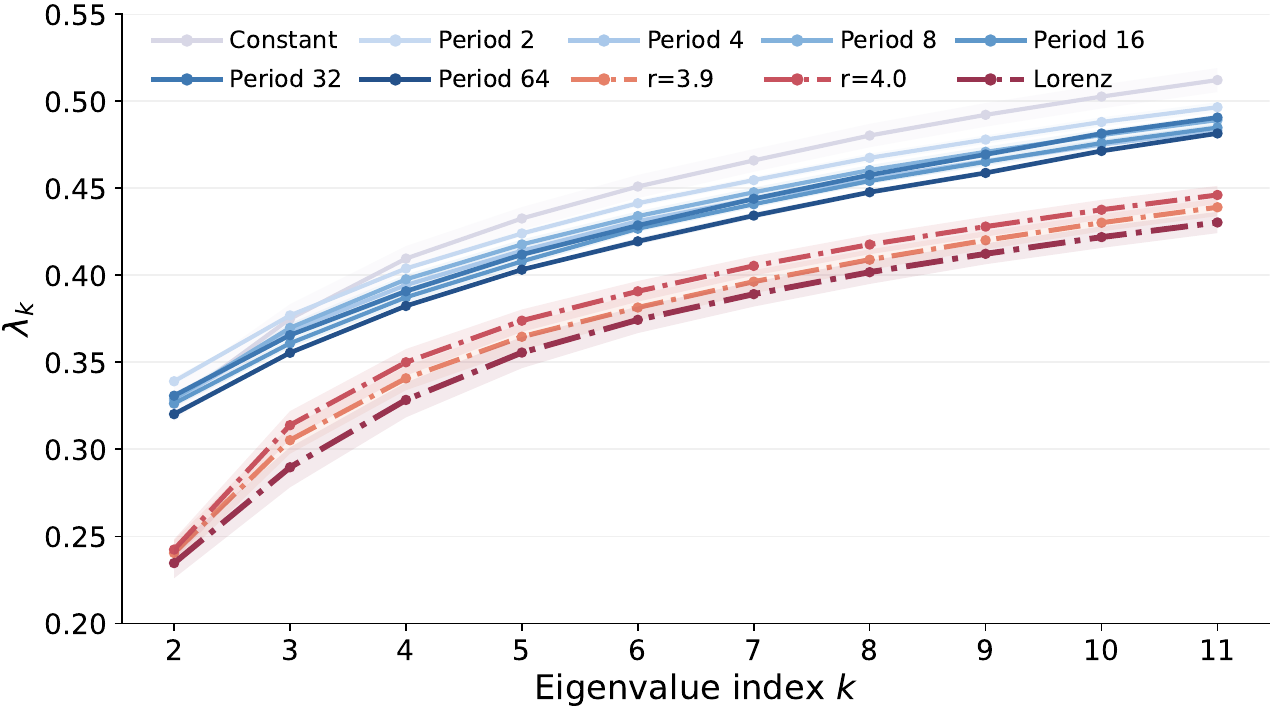}
  \caption{Low end of the normalized Laplacian spectrum at $N=2000$, averaged across layers. For each $k=2,\ldots,11$, curves show the mean $\lambda_k$ across 20 realizations, and shading denotes $\pm 1$ standard deviation. The chaotic logistic map and Lorenz inputs remain below the constant and periodic inputs throughout this range, showing that the separation observed in the normalized Fiedler value in \cref{fig:layer-averaged-combined-diagnostics} extends across the low end of the spectrum.}
  \label{fig:layer-averaged-low-end-laplacian-spectrum}
\end{figure}

\paragraph{Low-end normalized Laplacian spectrum.}
Let the ordered eigenvalues of the layer-$\ell$ normalized Laplacian $\mathcal L^{(\ell)}$ be
\[
0=\lambda_1^{(\ell)}
\leq \lambda_2^{(\ell)}
\leq \cdots
\leq \lambda_N^{(\ell)}
\leq 2.
\]
Writing $\lambda_k$ without a layer superscript for the layer average, we define
\[
\lambda_k
:=
\frac{1}{L}
\sum_{\ell=1}^{L}
\lambda_k^{(\ell)}.
\]
Thus, $\lambda_2$ is the layer-averaged normalized Fiedler value reported in the main analysis. To determine whether the observed separation extends beyond the two-way connectivity captured by $\lambda_2$, we examine the first ten nontrivial eigenvalues, $k=2,\ldots,11$. The additional eigenvalues at the low end of the normalized Laplacian spectrum are related to multiway sparse-partition structure through higher-order Cheeger inequalities \citep{lee2014multiway}. We focus on $N=2000$, where \cref{fig:layer-averaged-combined-diagnostics} in \Cref{sec:spectral-diagnostics} shows a clear separation in $\lambda_2$ between the chaotic and non-chaotic inputs. \cref{fig:layer-averaged-low-end-laplacian-spectrum} shows that this separation persists throughout the examined low end of the spectrum: the chaotic logistic map and Lorenz inputs remain systematically below the constant and periodic inputs for $k=2,\ldots,11$. This systematic downward shift is consistent with a greater tendency toward multiple weakly connected graph regions under chaotic inputs, indicating that the reduced global integration observed through $\lambda_2$ is not confined to a single eigenvalue.

\begin{figure*}
  \centering
  \includegraphics[width=\textwidth]{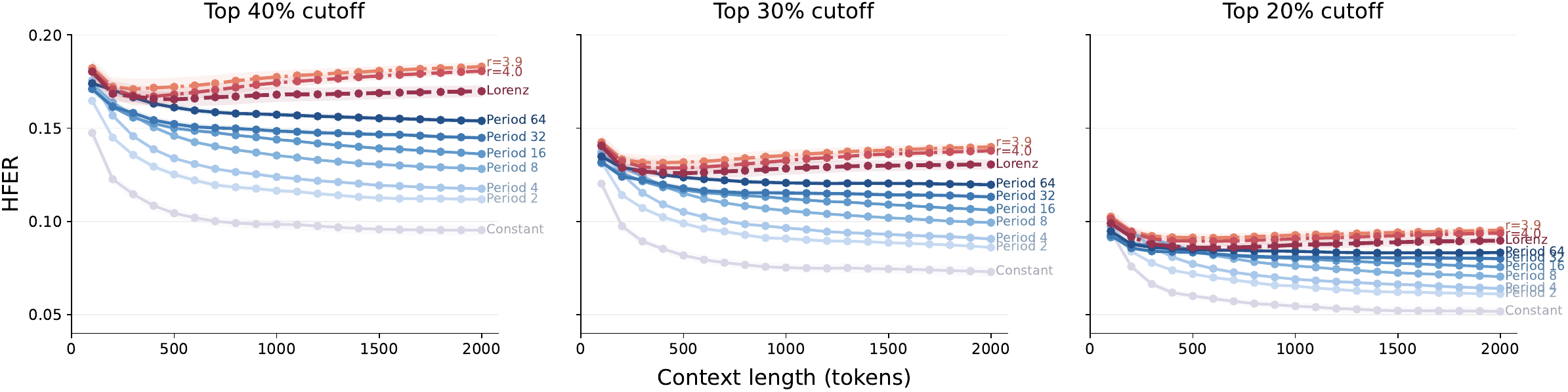}
  \caption{Layer-averaged HFER cutoff sensitivity. The panels report $\mathrm{HFER}_q$ using the highest-frequency 40\%, 30\%, and 20\% of graph Fourier modes. Under the same evaluation setup as \cref{sec:spectral-diagnostics}, curves and shading show the mean and $\pm 1$ standard deviation across 20 realizations. Across all three cutoffs, chaotic inputs exhibit higher HFER at longer contexts, while the non-chaotic inputs retain the periodicity-dependent ladder observed with the default 50\% high-frequency tail in \cref{fig:layer-averaged-combined-diagnostics}. The qualitative separation is therefore robust to the number of high-frequency modes included.}
  \label{fig:layer-averaged-hfer-cutoffs}
\end{figure*}

\begin{figure*}
  \centering
  \begin{subfigure}[t]{\textwidth}
    \centering
    \includegraphics[width=\linewidth]{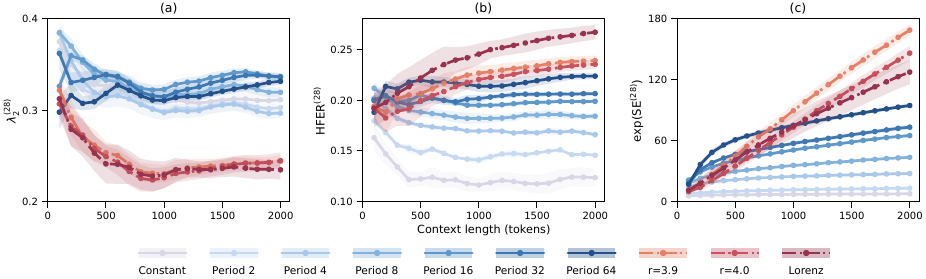}
    \captionsetup{skip=6pt}
    \caption*{(i) Final layer across context lengths}
    \vspace{8pt}
  \end{subfigure}
  \begin{subfigure}[t]{\textwidth}
    \centering
    \includegraphics[width=\linewidth]{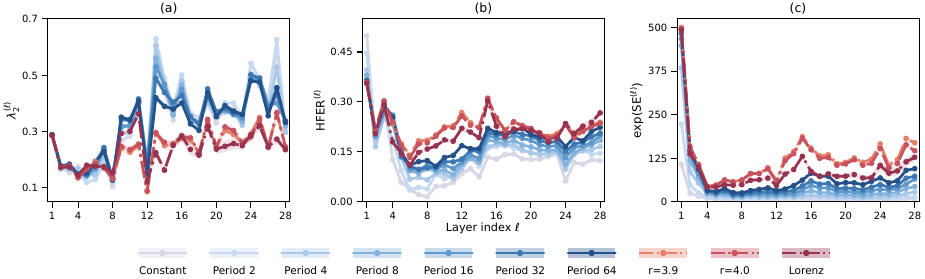}
    \captionsetup{skip=6pt}
    \caption*{(ii) All layers at $N=2000$}
  \end{subfigure}
  \caption{Layer-specific graph-spectral diagnostics for the main numerical input families using Llama-3.2-3B. Row (i) shows final-layer diagnostics across context lengths, while row (ii) shows the corresponding diagnostics across all 28 model layers at $N=2000$. Within each row, panels (a)--(c) report normalized Fiedler value $\lambda_2$, HFER, and effective spectral support $\exp(\mathrm{SE})$, respectively. Curves and shading denote the mean and $\pm 1$ standard deviation across the same 20 realizations analyzed in \cref{sec:spectral-diagnostics}. The final-layer results reproduce the broad context- and complexity-dependent behavior of the layer-averaged analysis, although the single-layer curves are less smooth. In the fixed-context depth profiles, most layers exhibit the same broad complexity-dependent ordering as the layer-averaged and final-layer results, with only a few of the earliest layers showing weaker separation.}
  \label{fig:layer-specific-combined-diagnostics}
\end{figure*}

\paragraph{HFER cutoff sensitivity.}
To test whether the observed HFER trends depend on the default 50\% high-frequency tail, let $
M_q=\operatorname{round}(qN)$ and $K_q=N-M_q$,
where $M_q$ is the number of highest-frequency modes retained for tail fraction $q$. We define
\begin{align*}
\mathrm{HFER}^{(\ell)}_q
&=
\sum_{m=K_q+1}^{N}
p_m^{(\ell)},\\
\mathrm{HFER}_q
&=
\frac{1}{L}
\sum_{\ell=1}^{L}
\mathrm{HFER}^{(\ell)}_q.
\end{align*}
Thus, $q=0.4$, $0.3$, and $0.2$ retain the highest-frequency 40\%, 30\%, and 20\% of graph Fourier modes, corresponding to cutoffs at $K_q=0.6N$, $0.7N$, and $0.8N$, respectively, for the evaluated context lengths. The default diagnostic in \Cref{sec:graph-spectral-diagnostics} is $\mathrm{HFER}_{0.5}$. As shown in \cref{fig:layer-averaged-combined-diagnostics} of \Cref{sec:spectral-diagnostics}, at longer contexts this default separates the higher-complexity chaotic inputs from the simpler constant and periodic inputs and reveals a clear ladder among the periodic inputs. \cref{fig:layer-averaged-hfer-cutoffs} shows the same qualitative behavior for $q=0.4$, $0.3$, and $0.2$: chaotic inputs retain higher HFER, while the periodic inputs preserve their period-dependent ordering. The persistence of these trends under increasingly selective high-frequency tails indicates that the observed behavior is not specific to the default 50\% high-frequency tail.

\subsection{Layerwise Graph-Spectral Diagnostics}
\label[appendix]{app:layer-resolved-spectral-diagnostics}
The main analysis in \Cref{sec:spectral-diagnostics} reports graph-spectral diagnostics averaged over transformer layers, providing a compact summary of how the diagnostic patterns vary across input families and context lengths. Here, we present two complementary views of the corresponding layerwise behavior. We first show the final-layer diagnostics across context lengths as an illustrative layer-specific example, and then examine all layers at $N=2000$ to characterize how the same broad input-family structure is distributed across model depth. Together, these views show that the layer-averaged results provide a holistic summary of the recurring layerwise behavior while smoothing layer-specific variation.

\paragraph{Final-layer context scaling.}
The top row of \cref{fig:layer-specific-combined-diagnostics} presents the three diagnostics at the final model layer as an illustrative layer-specific example. At longer contexts, the final-layer normalized Fiedler value primarily separates chaotic from non-chaotic inputs, whereas HFER and effective spectral support also reveal a finer period-dependent ordering among the periodic families. These patterns are broadly consistent with the layer-averaged analysis, although the single-layer curves are less smooth.

\paragraph{Consistency across model depth.}
Because presenting a separate context-length sweep for each layer of an $L$-layer model would require $L$ figures, we complement the final-layer sweep with a fixed-context depth profile that compares all layers in a single view. The bottom row of \cref{fig:layer-specific-combined-diagnostics} reports the normalized Fiedler value, HFER, and effective spectral support across all model layers at $N=2000$. Overall, the layerwise profiles are consistent with the layer-averaged and final-layer results, with most layers exhibiting the same broad complexity-dependent ordering.

\section{Attention-Induced Token Graphs: Construction and Visualization Robustness}
\label[appendix]{sec:attention-graph-construction}

\begin{figure*}[t]
  \centering
  \includegraphics[width=\textwidth]{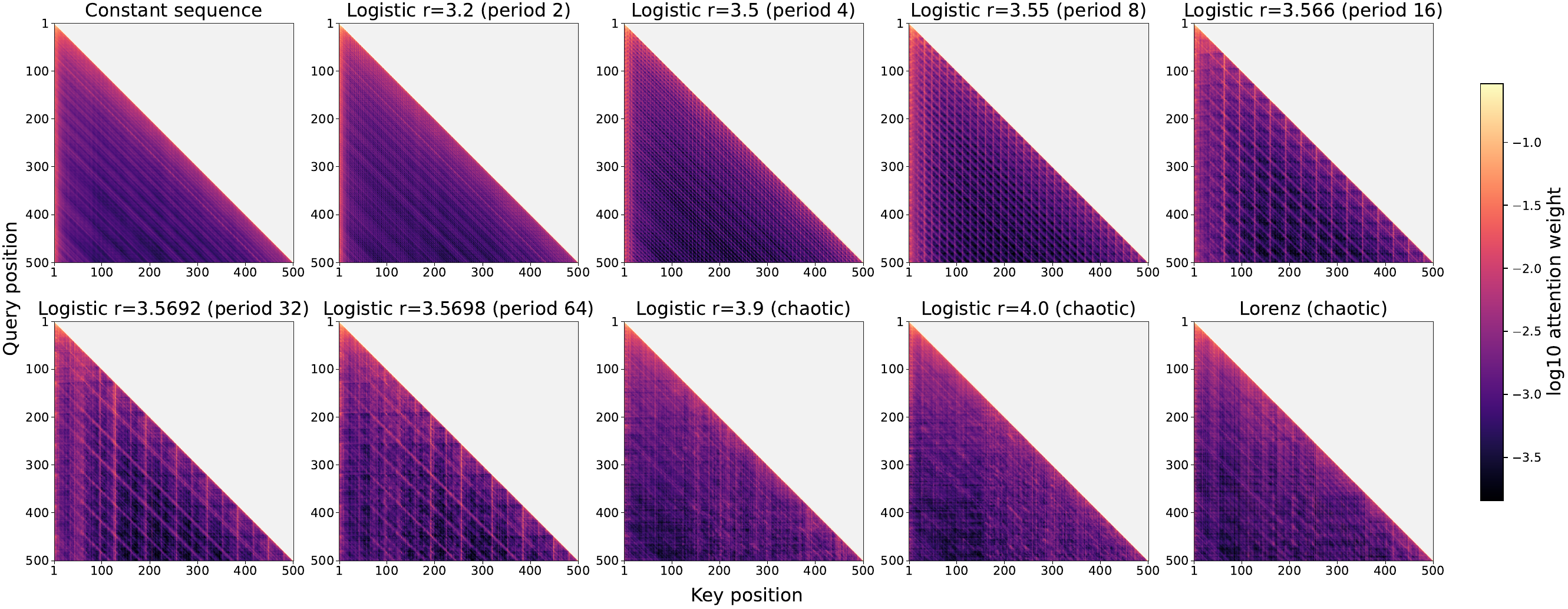}
  \caption{Last-layer attention matrices for Llama-3.2-3B across the main numerical input families at $N=500$. Each panel shows the head-averaged matrix for a representative input.}
  \label{fig:last-layer-attention-matrices}
\end{figure*}

\paragraph{Attention matrices.}
\cref{fig:last-layer-attention-matrices} shows representative final-layer, head-averaged post-softmax attention matrices for the main numerical input families. These matrices are symmetrized to form the weighted adjacency matrices used in the spectral diagnostics and graph visualizations.

\begin{figure*}
  \centering
  \includegraphics[width=\textwidth]{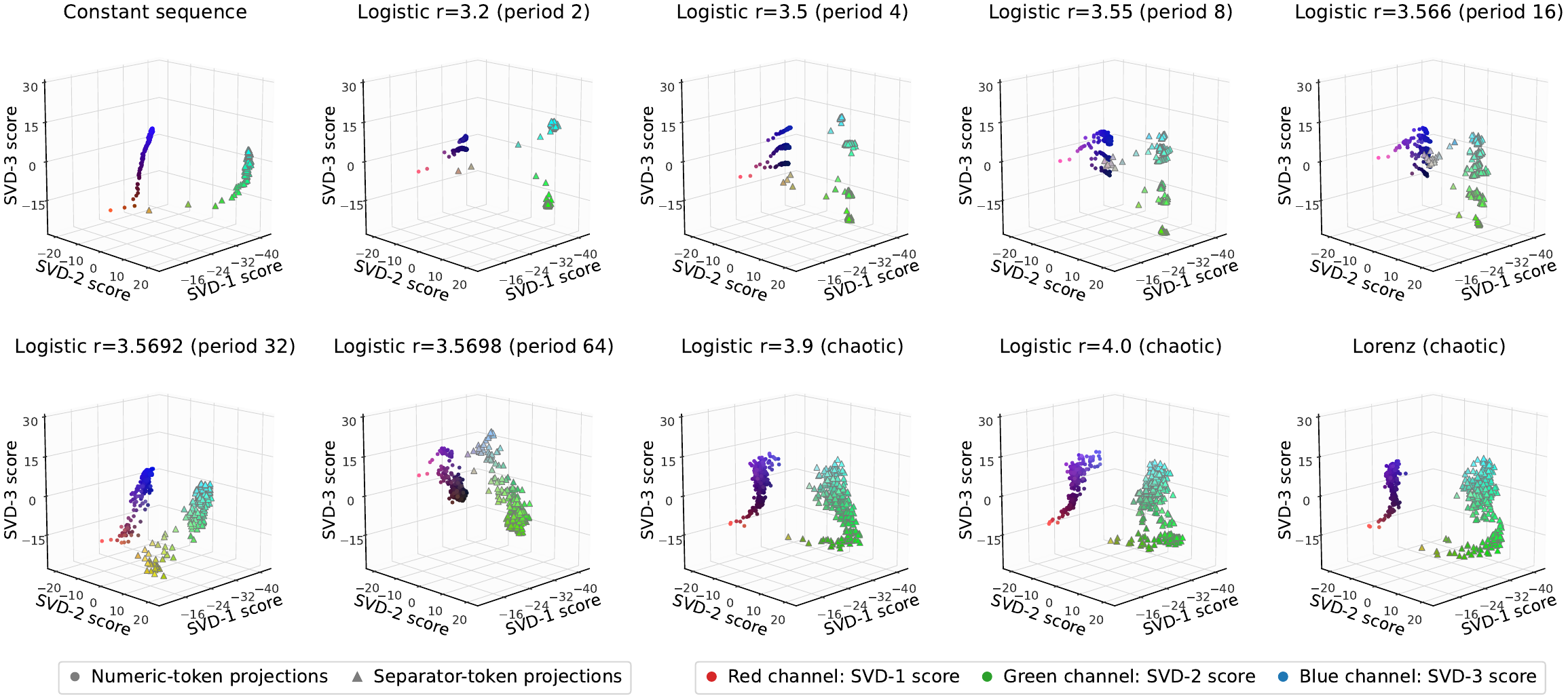}
  \caption{Last-layer hidden-state projections for Llama-3.2-3B across the main numerical input families at $N=500$. Each panel shows token hidden states projected onto the leading three uncentered SVD coordinates, which define the RGB node colors used in the attention-induced graph visualizations.}
  \label{fig:last-layer-signal-visualization-all-inputs}
\end{figure*}

\begin{figure}
  \centering
  \includegraphics[width=0.75\columnwidth]{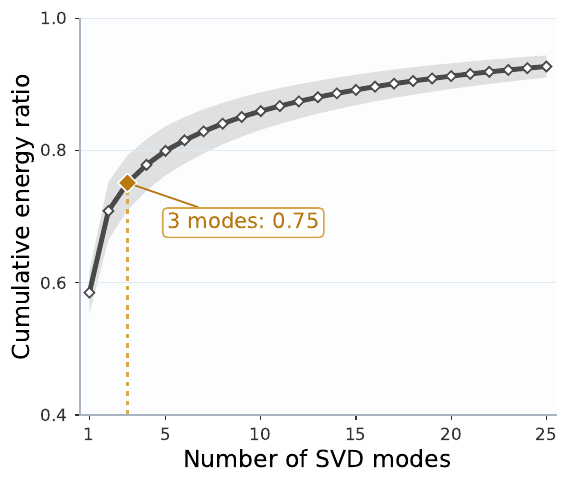}
  \caption{Cumulative hidden-state SVD-energy fraction for Llama-3.2-3B at $N=500$, averaged first across the main numerical input families within each layer and then across the analyzed layers. The leading three components capture 75\% of the hidden-state energy on average, supporting their use as RGB coordinates for token-node colors.}
  \label{fig:cumulative-svd-energy}
\end{figure}

\paragraph{Hidden-state projections.}
Let $X^{(\ell)}\in\mathbb{R}^{N\times d}$ contain the token hidden states at layer $\ell$, with one token per row. We compute the uncentered singular value decomposition $X^{(\ell)}
=
U_X^{(\ell)}
\Sigma_X^{(\ell)}
V_X^{(\ell)\top}$,
where the subscript $X$ distinguishes these hidden-state SVD quantities from the graph Laplacian eigenvectors $U^{(\ell)}$ introduced in \Cref{sec:graph-spectral-diagnostics}. The full matrix of token-level SVD coordinates is
\[
Z^{(\ell)}
=
U_X^{(\ell)}\Sigma_X^{(\ell)}
=
X^{(\ell)}V_X^{(\ell)}.
\]
To construct the node colors, we assign the first three columns of $Z^{(\ell)}$ to the red, green, and blue channels, respectively. For each $k\in\{1,2,3\}$, let $m_k^{(\ell)}=\min_j Z_{jk}^{(\ell)}$ and $M_k^{(\ell)}=\max_j Z_{jk}^{(\ell)}$. Using a numerical tolerance of $\varepsilon=10^{-12}$, we independently min--max scale each coordinate across nodes:
\[
C_{ik}^{(\ell)}
=
\begin{cases}
\displaystyle
\frac{Z_{ik}^{(\ell)}-m_k^{(\ell)}}
{M_k^{(\ell)}-m_k^{(\ell)}},
& M_k^{(\ell)}-m_k^{(\ell)}>\varepsilon,\\[9pt]
1/2,
& \text{otherwise}.
\end{cases}
\]
Token $i$ is then assigned the RGB color $\mathbf c_i^{(\ell)}=(C_{ik}^{(\ell)})_{k=1}^{3}\in[0,1]^3$. The fraction of hidden-state energy captured by the first $q$ components is
\[
\eta_q^{(\ell)}
\,=\,
\frac{\displaystyle\sum_{k=1}^{q}\bigl(\sigma_{X,k}^{(\ell)}\bigr)^2}
{\displaystyle\sum_{k=1}^{r}\bigl(\sigma_{X,k}^{(\ell)}\bigr)^2}
\,=\,
\frac{\displaystyle\sum_{k=1}^{q}\bigl(\sigma_{X,k}^{(\ell)}\bigr)^2}
{\lVert X^{(\ell)}\rVert_F^2}.
\]
Here, $r=\min(N,d)$ is the number of singular values.
At $N=500$, averaging first across the main numerical input families within each layer and then across the analyzed layers gives $\eta_3=0.75$. \cref{fig:last-layer-signal-visualization-all-inputs} shows the resulting final-layer projections for all ten families, while \cref{fig:cumulative-svd-energy} includes the layer-averaged cumulative SVD-energy diagnostic and its variation across layers.

\begin{figure*}
  \centering
  \includegraphics[width=\textwidth]{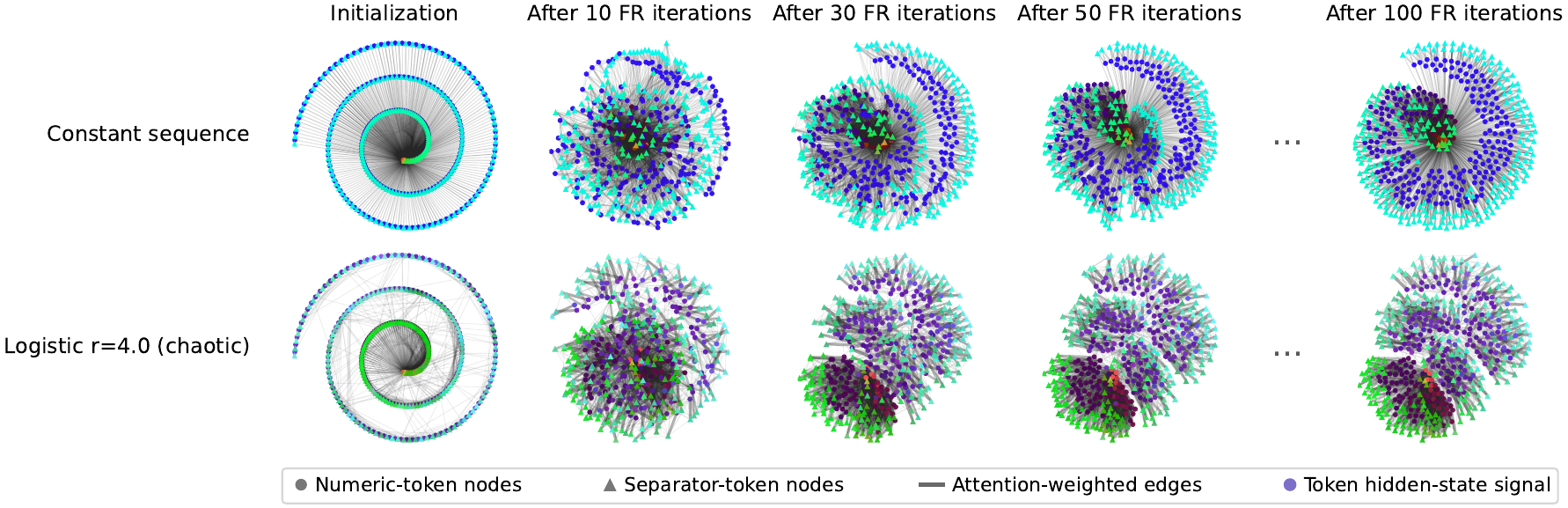}
  \caption{Evolution of attention-based node positions for Llama-3.2-3B at $N=500$. Representative constant and chaotic logistic map inputs are shown at initialization and after 10, 30, 50, and 100 FR iterations. Both begin from the same token-ordered spiral, so subsequent differences reflect their input-specific attention weights. Most visible reorganization occurs by 50 iterations, with only minor changes thereafter; we use the 100-iteration layouts for subsequent token-graph visualizations.}
  \label{fig:last-layer-fr-iteration-trajectory}
\end{figure*}

\paragraph{Layout computation.}
We initialize node $i$ on a token-ordered spiral by setting $t_i
=(i-1)/(N-1)$, $\theta_i=-\pi/2+2\pi K_Nt_i$, and 
\[
\mathbf p_i
=
(0.15+0.85t_i)
\begin{bmatrix}
\cos\theta_i\\
\sin\theta_i
\end{bmatrix},
\]
where $K_N=2.25$ for $N<80$ and $K_N=2.75$ otherwise. We center the resulting coordinates and rescale them to unit maximum radius. This construction produces the initialization shown in the leftmost panels of \cref{fig:last-layer-fr-iteration-trajectory}. Because it depends only on $N$ and token order, all inputs of the same length begin from identical node positions; their input-specific attention matrices $\bar W^{(\ell)}$ then determine the attractive forces that drive the subsequent reorganization. We compute the resulting positions using the implementation of the FR algorithm provided by the NetworkX Python library \citep{fruchterman1991graph,hagberg2008exploring}, with all off-diagonal entries of $\bar W^{(\ell)}$ used as edge weights. We run 100 iterations rather than the implementation's default of 50 to allow further relaxation toward a stable configuration.

\paragraph{Layout evolution and iteration count.}
\Cref{fig:last-layer-fr-iteration-trajectory} shows the evolution of the FR layout from the shared token-ordered spiral for representative constant and chaotic logistic map inputs. Most visible layout reorganization occurs by 50 iterations, with only minor changes thereafter. We therefore use the 100-iteration layouts throughout the token-graph visualizations.

\begin{figure}
  \centering
  \includegraphics[width=\columnwidth]{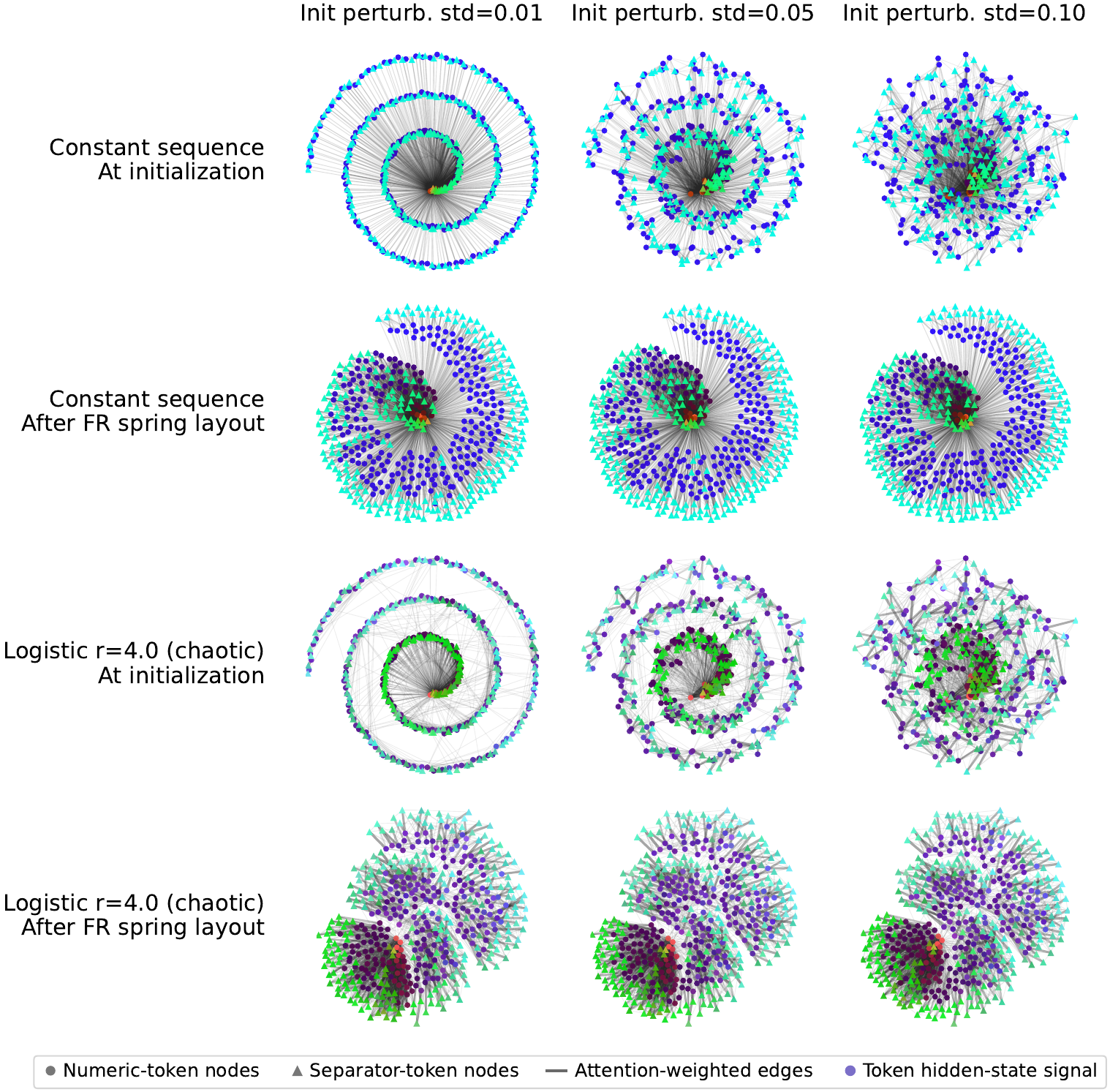}
  \caption{Layout-initialization robustness of last-layer attention-graph visualizations for Llama-3.2-3B at $N=500$. Constant and chaotic logistic map examples are initialized from noisy spiral coordinates with perturbation standard deviations $0.01$, $0.05$, and $0.10$, then compared before and after FR layout relaxation. The qualitative organization is preserved across perturbation scales.}
  \label{fig:last-layer-noisy-spiral-initialization-robustness}
\end{figure}

\paragraph{Initialization robustness.}
To evaluate sensitivity to the exact initialization, we add independent Gaussian perturbations $\boldsymbol{\epsilon}_i\sim\mathcal N(\mathbf 0,\sigma^2I_2)$ to the normalized spiral coordinates, reapply the same normalization, and rerun the 100-iteration layout. \cref{fig:last-layer-noisy-spiral-initialization-robustness} compares $\sigma\in\{0.01,0.05,0.10\}$ for representative constant and chaotic logistic map inputs. Their broad organization remains stable across these perturbation magnitudes, indicating that the qualitative separation is not specific to the initial spiral coordinates.

\begin{figure}
  \centering
  \includegraphics[width=\columnwidth]{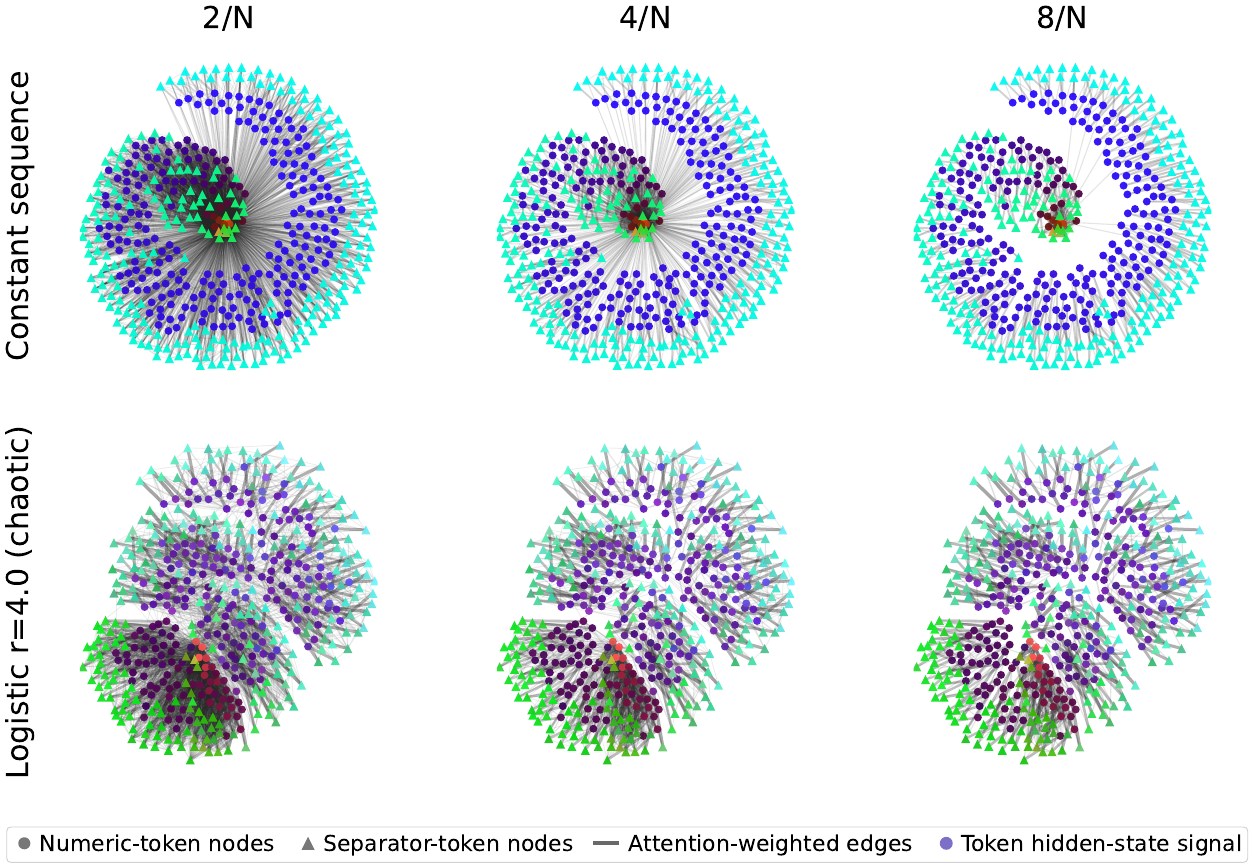}
  \caption{Threshold robustness of last-layer attention-graph visualizations for Llama-3.2-3B at $N=500$. Representative constant and chaotic logistic map inputs are shown for $\kappa\in\{2,4,8\}$, with display threshold $\kappa/N$. The main visualizations use $\kappa=2.5$. The qualitative structure remains stable across thresholds.}
  \label{fig:last-layer-graph-threshold-robustness}
\end{figure}

\paragraph{Edge-display threshold sensitivity.}
\cref{fig:last-layer-graph-threshold-robustness} evaluates sensitivity to the multiplier $\kappa$ in the edge-display threshold $\bar W^{(\ell)}_{ij}>\kappa/N$. Because this threshold affects only edge rendering, node positions, hidden-state colors, and graph-spectral diagnostics remain unchanged across panels. For both the representative constant and chaotic logistic map inputs, the qualitative structure of the displayed edge patterns remains consistent across $\kappa\in\{2,4,8\}$ and with the default visualization at $\kappa=2.5$.

\section{Attention-Induced Token-Graph Visualizations at Additional Transformer Layers}
\label[appendix]{sec:token_graphs}

\begin{figure*}
  \centering
  \begin{subfigure}[t]{\textwidth}
    \centering
    \includegraphics[width=0.95\linewidth]{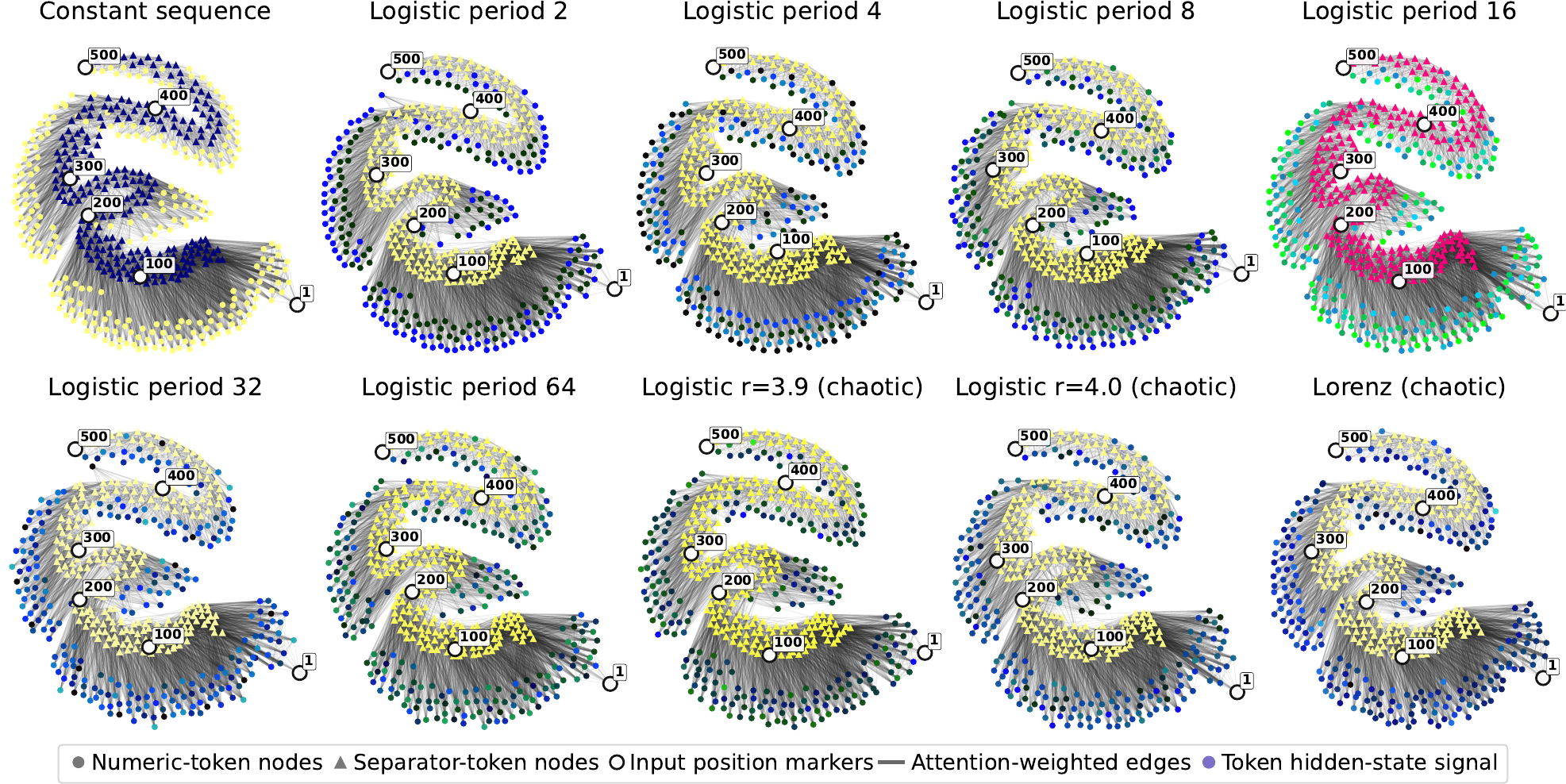}
    \caption{Transformer layer 1}
  \end{subfigure}
  \par\vspace{2em}
  \begin{subfigure}[t]{\textwidth}
    \centering
    \includegraphics[width=0.95\linewidth]{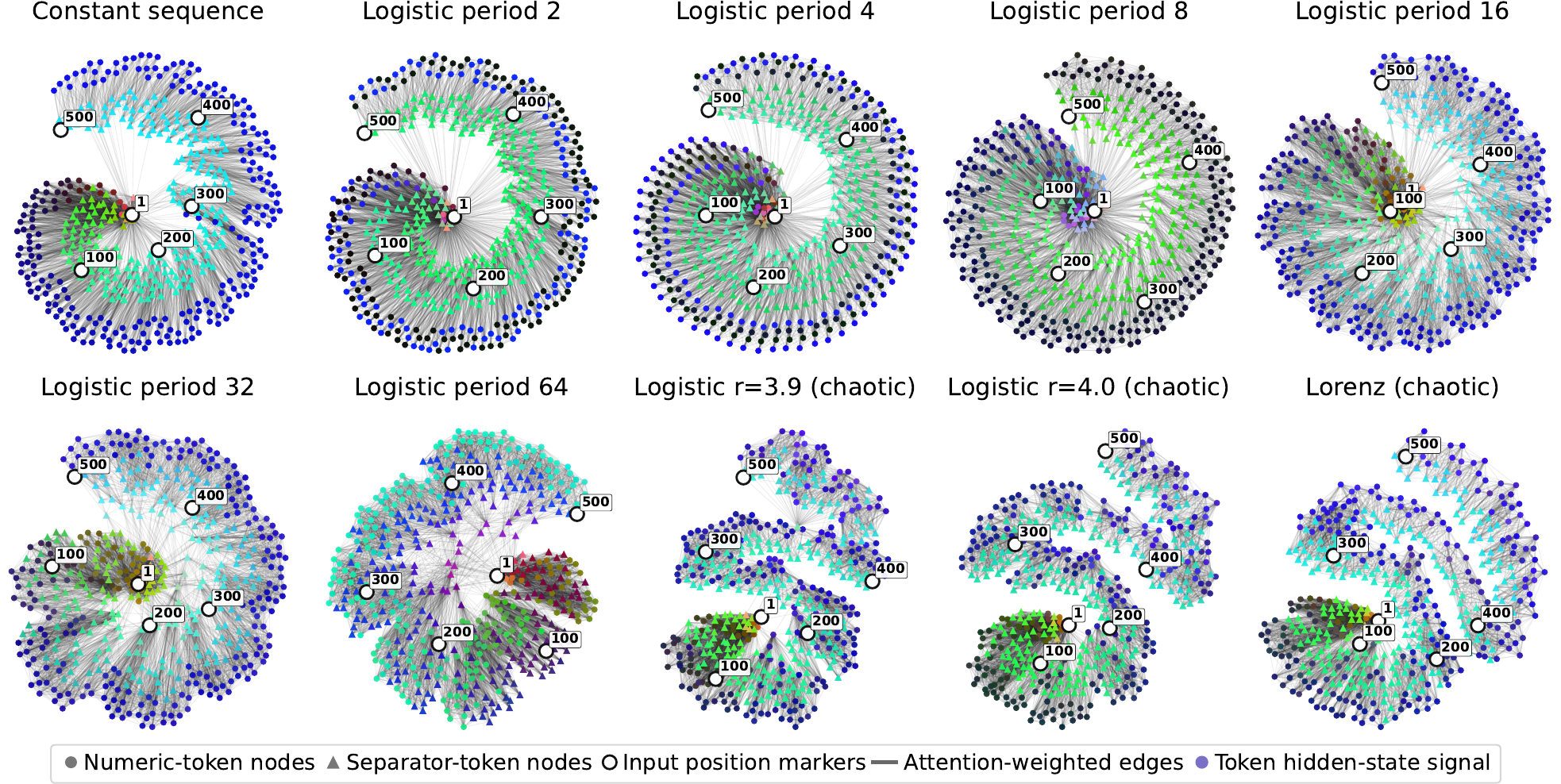}
    \caption{Transformer layer 12}
  \end{subfigure}
  \caption{Layerwise attention-induced token graphs at $N=500$ for the main numerical input families examined in \Cref{sec:attention-graph-visualizations}. \textbf{(a)} Layer 1 provides an illustrative early-depth snapshot in which cross-input differentiation in graph layout is comparatively limited and the layouts retain broadly similar local organization. \textbf{(b)} Layer 12 provides an illustrative middle-depth snapshot in which cross-input differentiation is more pronounced, with graph organization tending to become less uniform as input dynamical complexity increases.}
  \label{fig:layer-wise-graphs}
\end{figure*}

\Cref{sec:attention-graph-visualizations} focuses on last-layer attention graphs because the final layer is closest to the prediction-facing representation used by the model. \cref{fig:layer-wise-graphs} uses layers 1 and 12 as illustrative snapshots of early and middle depth, respectively. In the layer-1 example, differentiation across input families is comparatively limited: although the graphs are not identical, their layouts share broad organization associated with numeric-versus-separator token roles and local sequence structure. In the layer-12 example, input-dependent differences are more pronounced and broadly track dynamical complexity: constant and lower-period inputs retain relatively regular, globally integrated layouts, whereas higher-period, chaotic logistic map, and Lorenz inputs exhibit less homogeneous organization and more pronounced mesoscale partitioning into localized subnetworks. These two layer-specific snapshots illustrate how cross-input differentiation can become more pronounced with depth; they do not establish a universal transition point or imply that all early and middle layers follow the same progression.

\section{Graph-Spectral Diagnostics Across Model Families, Scales, and Instruction Tuning}
\label[appendix]{app:model-scale-instruction-tuning}

We investigate whether the graph-spectral behavior observed in the main analysis persists across model families and scales, and how instruction tuning affects this behavior. We first examine variation across model scale and instruction tuning within the Llama 3 family. We then extend the analysis to Phi-4 and SmolLM3, including a comparison between the base and instruction-tuned SmolLM3 variants.

\begin{figure*}
  \centering
  \includegraphics[width=\textwidth]{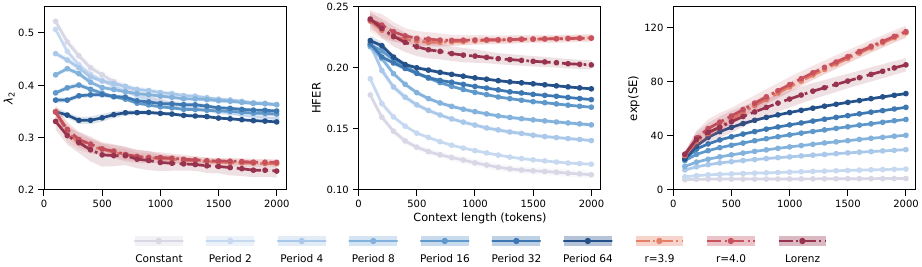}
  {\captionsetup{skip=2pt}\caption*{(a) Llama-3.1-8B}}
  \includegraphics[width=\textwidth]{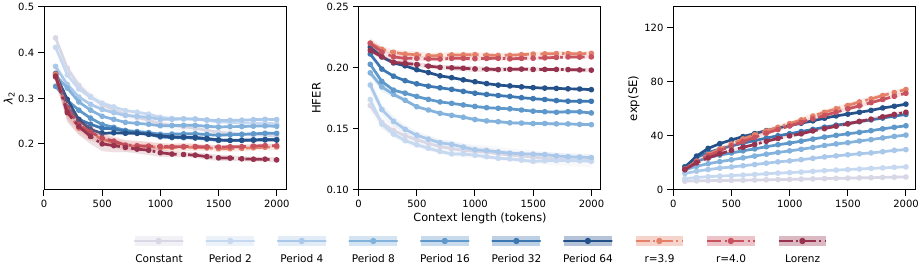}
  {\captionsetup{skip=2pt}\caption*{(b) Llama-3.2-1B}}
  \includegraphics[width=\textwidth]{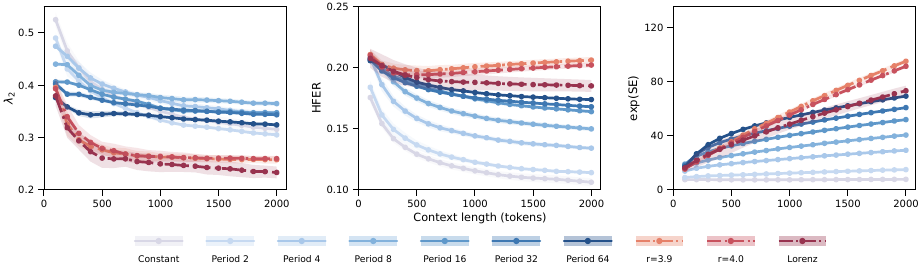}
  {\captionsetup{skip=2pt}\caption*{(c) Llama-3.1-8B-Instruct}}
  \includegraphics[width=\textwidth]{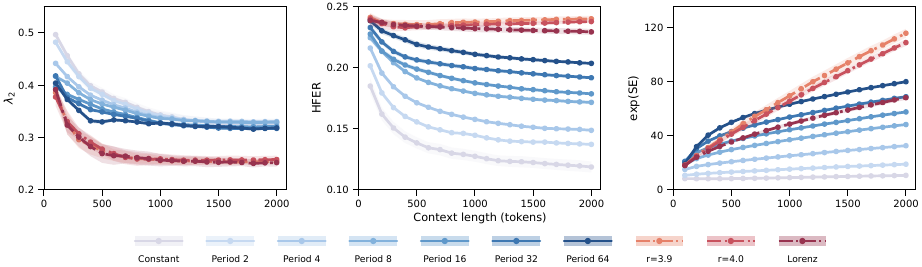}
  {\captionsetup{skip=2pt}\caption*{(d) Llama-3.2-3B-Instruct}}
  \caption{Layer-averaged graph-spectral diagnostics across base-model scales and instruction-tuned variants within the Llama 3 family, under the same experimental setup as \Cref{sec:spectral-diagnostics}. From left to right, the columns report normalized Fiedler value $\lambda_2$, HFER, and $\exp(\mathrm{SE})$ as functions of context length. Compared with the Llama-3.2-3B results in \cref{fig:layer-averaged-combined-diagnostics}, the 8B base model exhibits highly similar behavior, whereas the 1B base model less clearly distinguishes the dynamical regimes. Both instruction-tuned models retain the broad behavior of their base counterparts, but some finer distinctions among input families become less pronounced with instruction tuning.}
  \label{fig:llama-family-layer-averaged-combined-diagnostics}
\end{figure*}

\paragraph{Variation across model scales within the Llama 3 family.}
Using the same experimental setup as \Cref{sec:spectral-diagnostics}, \cref{fig:llama-family-layer-averaged-combined-diagnostics}(a)--(b) compares the layer-averaged diagnostics for Llama-3.2-1B and Llama-3.1-8B with the primary Llama-3.2-3B results. The 8B model exhibits behavior highly similar to that of the 3B model, preserving the broad trends and separation among input families across the three diagnostics. The 1B model, however, distinguishes the input families less clearly. In particular, its $\lambda_2$ and $\exp(\mathrm{SE})$ results show less clear separation between the chaotic and high-period periodic inputs, while its HFER exhibits a less distinct complexity-ordered progression among the lower-period inputs. These results suggest that the smaller 1B model retains the broad context-dependent graph-spectral trends but less consistently resolves fine-grained differences between dynamical regimes.

\paragraph{Instruction-tuned model comparison.}
We repeat the experimental setup from \Cref{sec:spectral-diagnostics} with the instruction-tuned variants of the 3B model used in the main analysis and the 8B model evaluated in the preceding scale comparison: Llama-3.2-3B-Instruct and Llama-3.1-8B-Instruct. Llama-3 Instruct models undergo alignment-focused post-training to support helpful and safe instruction following \citep{grattafiori2024llama3}. We focus on the 3B and 8B models because the preceding scale comparison shows weaker input-family separation for the 1B base model. As shown in \cref{fig:llama-family-layer-averaged-combined-diagnostics}(c)--(d), both instruction-tuned models retain the principal context- and complexity-dependent behavior of their base counterparts, although some fine-grained distinctions between high-period periodic and chaotic inputs become less pronounced, particularly in $\exp(\mathrm{SE})$. This weakening is consistent with output-level evidence that alignment-related post-training can reduce extrapolation accuracy for time series \citep{gruver2024largelanguagemodelszeroshot} and spatiotemporal dynamics \citep{bao2026texttrained}.

\begin{figure*}
  \centering
  \includegraphics[width=\textwidth]{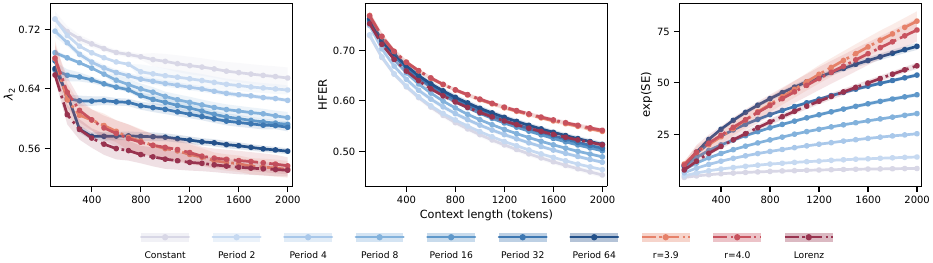}
  {\captionsetup{skip=2pt}\caption*{(a) Phi-4}}
  \vspace{3pt}
  \includegraphics[width=\textwidth]{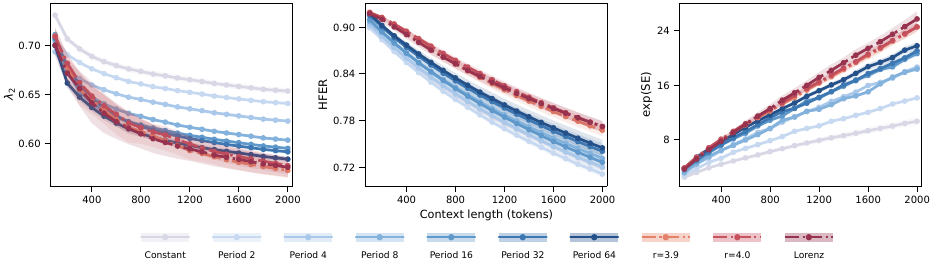}
  {\captionsetup{skip=2pt}\caption*{(b) SmolLM3-3B-Base}}
  \vspace{3pt}
  \includegraphics[width=\textwidth]{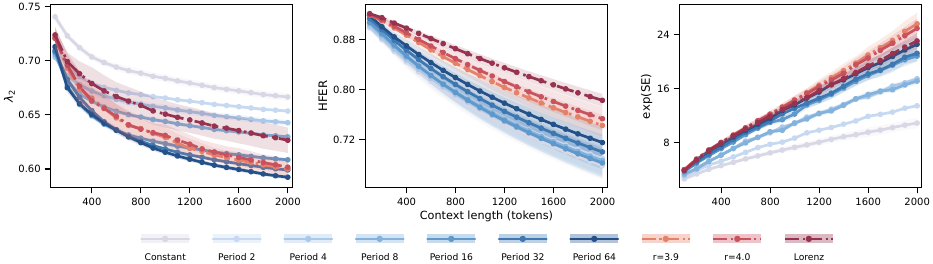}
  {\captionsetup{skip=2pt}\caption*{(c) SmolLM3-3B (instruction-tuned)}}
  \caption{Layer-averaged graph-spectral diagnostics for additional model families under the same experimental setup as \Cref{sec:spectral-diagnostics}. From left to right, the columns report normalized Fiedler value $\lambda_2$, HFER, and $\exp(\mathrm{SE})$ as functions of context length. SmolLM3-3B-Base reproduces the broad context- and complexity-dependent behavior observed in the Llama 3 models. Phi-4, a post-trained chat model, and instruction-tuned SmolLM3-3B also retain the overall trends but exhibit less consistent fine-grained separation among high-period and chaotic inputs. Diagnostic magnitudes and rates of change with context length differ across models.}
  \label{fig:additional-model-families-layer-averaged-combined-diagnostics}
\end{figure*}

\paragraph{Additional model families.}
Using the same experimental setup as \Cref{sec:spectral-diagnostics}, \cref{fig:additional-model-families-layer-averaged-combined-diagnostics} presents the layer-averaged diagnostics for Phi-4 and the base and instruction-tuned SmolLM3-3B variants. At the time of our experiments, only the post-trained, chat-optimized Phi-4 checkpoint was publicly available; unlike Llama 3 and SmolLM3, Phi-4 had no corresponding publicly released base-model checkpoint. Across all three models, the broad context- and complexity-dependent behavior remains consistent with the Llama 3 results. SmolLM3-3B-Base closely reproduces the ordered input-family behavior of the 3B and 8B Llama base models, whereas Phi-4 and instruction-tuned SmolLM3-3B show less clear separation between high-period periodic and chaotic inputs, similar to the instruction-tuned Llama variants. The absolute diagnostic magnitudes and rates of change with context differ across models, indicating that the detailed graph-spectral signatures are model-dependent even though their overall ordering and trends remain broadly consistent.

\section{Attention-Only and Hidden-State-Only Non-Graph Baselines}
\label[appendix]{app:non-graph-controls}

We further show that the broad context- and complexity-dependent trends captured by the graph-spectral diagnostics are not artifacts of the graph-signal construction. Analogous summaries derived from attention alone or hidden states alone exhibit similar behavior as context length increases and across input families, although they provide less fine-grained separation among families. As non-graph controls, we consider two hidden-state-only summaries based on fixed frequencies over serialized token position and one attention-only summary based on attention concentration. Following the notation of \Cref{sec:graph-spectral-diagnostics}, let $A^{(\ell,h)}\in\mathbb R^{N\times N}$ denote the post-softmax attention matrix for layer $\ell$ and head $h$, and let $X^{(\ell)}\in\mathbb R^{N\times d}$ denote the hidden-state matrix entering layer $\ell$. The positional controls use only $X^{(\ell)}$, whereas the attention control uses only $A^{(\ell,h)}$.

\paragraph{Normalized attention Gini coefficient.}
The Gini coefficient measures the concentration of a nonnegative vector \citep{hurley2009comparing}. For a nonzero vector $\mathbf z\in\mathbb R_{\geq 0}^m$, define
\[
g(\mathbf z)
=
\frac{\sum_{r=1}^{m}\sum_{s=1}^{m}|z_r-z_s|}
{2m\sum_{r=1}^{m}z_r}.
\]
This coefficient satisfies $g(\mathbf z)\in[0,1-1/m]$: it equals $0$ if and only if the entries of $\mathbf z$ are uniform, and reaches $1-1/m$ when all mass is concentrated on a single entry. For query position $i$ and head $h$, let $\mathbf a_i^{(\ell,h)}=(A_{i1}^{(\ell,h)},\ldots,A_{ii}^{(\ell,h)})$ contain its causally available attention weights. We normalize by the maximum attainable for each row length and average across query positions and heads:
\[
\mathrm{Gini}^{(\ell)}
=
\frac{1}{H(N-1)}
\sum_{h=1}^{H}\sum_{i=2}^{N}
\frac{g\!\left(\mathbf a_i^{(\ell,h)}\right)}{1-1/i}.
\]
The normalization makes rows with different numbers of available keys comparable and gives $\mathrm{Gini}^{(\ell)}\in[0,1]$; the first position is omitted because it has only one available key. A value of $0$ means that every included query distributes attention uniformly over its available keys, whereas a value of $1$ means that every query concentrates all attention on a single key. Intermediate values quantify the average degree of attention concentration. In this sense, attention Gini is an attention-only control related to the normalized Fiedler value $\lambda_2$ from our graph-spectral diagnostics: greater rowwise concentration may accompany the more localized attention-graph structure indicated by smaller $\lambda_2$, but Gini does not measure how those attention links combine into a globally connected graph.

\paragraph{Fixed positional-frequency representation.}
The two hidden-state controls measure how hidden states vary across token positions in the serialized input, without using attention. We construct this positional-frequency representation using an orthonormal type-II discrete cosine transform (DCT-II) \citep{ahmed1974discrete}. Unlike a standard discrete Fourier transform, the DCT-II does not impose periodic wraparound between the first and last token positions, while its orthonormality preserves total squared hidden-state energy. In implementation, we apply SciPy's \texttt{scipy.fft.dct} with \texttt{type=2}, \texttt{norm="ortho"}, and \texttt{axis=0}, transforming the token dimension independently for each hidden-state coordinate. We write the resulting coefficients as
\[
\widetilde X_{\mathrm{pos}}^{(\ell)}
=
\operatorname{DCT}_{\mathrm{II}}\!\left(X^{(\ell)}\right).
\]
Let $\widetilde{\mathbf x}_{\mathrm{pos},k}^{(\ell)}\in\mathbb R^d$ denote the coefficient vector for positional mode $k$. These modes are ordered from slow to rapid variation in token hidden states across serialized positions, rather than by temporal frequency in the underlying numerical trajectory. The fixed DCT basis thus provides a hidden-state-only counterpart to the graph Fourier basis used in our main diagnostics: both decompose hidden states into frequency-ordered modes, but the DCT basis is determined only by context length and serialized token order, whereas the graph Fourier basis is determined by the attention-induced token geometry.

\paragraph{Positional HFER.}
Let $p_{\mathrm{pos},k}^{(\ell)}=\|\widetilde{\mathbf x}_{\mathrm{pos},k}^{(\ell)}\|_2^2/\sum_{r=1}^{N}\|\widetilde{\mathbf x}_{\mathrm{pos},r}^{(\ell)}\|_2^2$ denote the normalized positional-modal energy. Using the same default cutoff $K=\lfloor N/2\rfloor$ as graph HFER, we define positional HFER as the fraction of hidden-state energy assigned to positional-frequency modes above the cutoff:
\[
\mathrm{HFER}_{\mathrm{pos}}^{(\ell)}
=
\sum_{k=K+1}^{N}p_{\mathrm{pos},k}^{(\ell)}.
\]
This quantity lies in $[0,1]$: it is $0$ when all energy lies in modes at or below the cutoff and $1$ when all energy lies in modes above the cutoff. Larger values indicate that a greater fraction of hidden-state energy is associated with rapid variation across serialized token positions. By comparison, graph HFER quantifies hidden-state variation over the attention-induced token geometry.

\paragraph{Positional spectral entropy (SE).}
Analogously to graph SE, positional SE measures how broadly hidden-state energy is distributed across the positional-frequency modes. Using the normalized positional-modal energies $p_{\mathrm{pos},k}^{(\ell)}$, we define
\[
\mathrm{SE}_{\mathrm{pos}}^{(\ell)}
=
-\sum_{k=1}^{N}
p_{\mathrm{pos},k}^{(\ell)}
\log p_{\mathrm{pos},k}^{(\ell)}.
\]
Following the main analysis, we report $\exp(\mathrm{SE}_{\mathrm{pos}}^{(\ell)})$ as the effective number of contributing positional modes. This quantity lies in $[1,N]$: it equals $1$ when all energy is concentrated in one mode and $N$ when energy is distributed uniformly across all modes. Larger values indicate that hidden-state energy is distributed across a broader set of positional-frequency modes. Thus, $\exp(\mathrm{SE}_{\mathrm{pos}}^{(\ell)})$ and graph-spectral $\exp(\mathrm{SE}^{(\ell)})$ quantify analogous notions of spectral breadth, but in the fixed positional DCT basis and the attention-induced graph Fourier basis, respectively.

\begin{figure}
  \centering
  \includegraphics[width=\columnwidth]{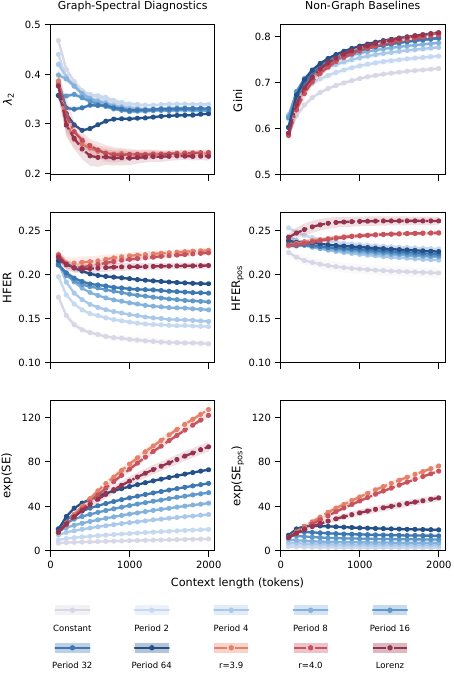}
  \caption{Comparison of graph-spectral diagnostics (left) and corresponding non-graph controls (right) across context lengths under the same evaluation setup as \Cref{sec:spectral-diagnostics}. The controls reproduce the broad context- and complexity-dependent trends. The top-row control is attention-only: increasing attention Gini indicates more concentrated attention. The lower two controls are hidden-state-only and measure high-frequency energy and effective spectral support in a fixed positional-frequency basis. The positional controls and corresponding graph-spectral diagnostics capture analogous spectral properties, but the latter show clearer separation among input families on matched y-axis scales, highlighting the added value of jointly analyzing hidden-state variation and attention-induced graph structure.}
  \label{fig:non-graph-baseline-comparison}
\end{figure}

\paragraph{Comparison with graph-spectral diagnostics.}
Under the same evaluation setup as \Cref{sec:spectral-diagnostics}, \cref{fig:non-graph-baseline-comparison} compares the three non-graph controls, each computed at every transformer layer and then averaged across the $L$ layers, with the main graph-spectral diagnostics across context lengths. These controls recover the broad context- and complexity-dependent trends across the main numerical input families, showing that the broad qualitative findings are also visible in attention- or hidden-state-only summaries and are therefore not artifacts of graph construction. When plotted with matched y-axis limits, positional HFER and $\exp(\mathrm{SE}_{\mathrm{pos}})$ exhibit narrower dynamic ranges and less separation among several input families. One notable difference between positional and graph HFER is that the former assigns the period-2 family greater high-frequency energy than periods 4--64 at every context length, reflecting the sensitivity of fixed positional modes to rapid alternation in serialized token order. Graph HFER instead measures hidden-state variation relative to the attention-induced token geometry, incorporating attention structure rather than relying on token order alone. The graph-spectral formulation therefore provides a joint characterization of attention-induced token connectivity and hidden-state variation and more consistently resolves the finer family- and context-specific behaviors discussed in \Cref{sec:spectral-diagnostics}.

\section{Additional Attention-Induced Token Graph Visualizations}
\label[appendix]{app:additional-token-graph-visualizations}

\begin{figure}
  \centering
  \includegraphics[width=0.95\columnwidth]{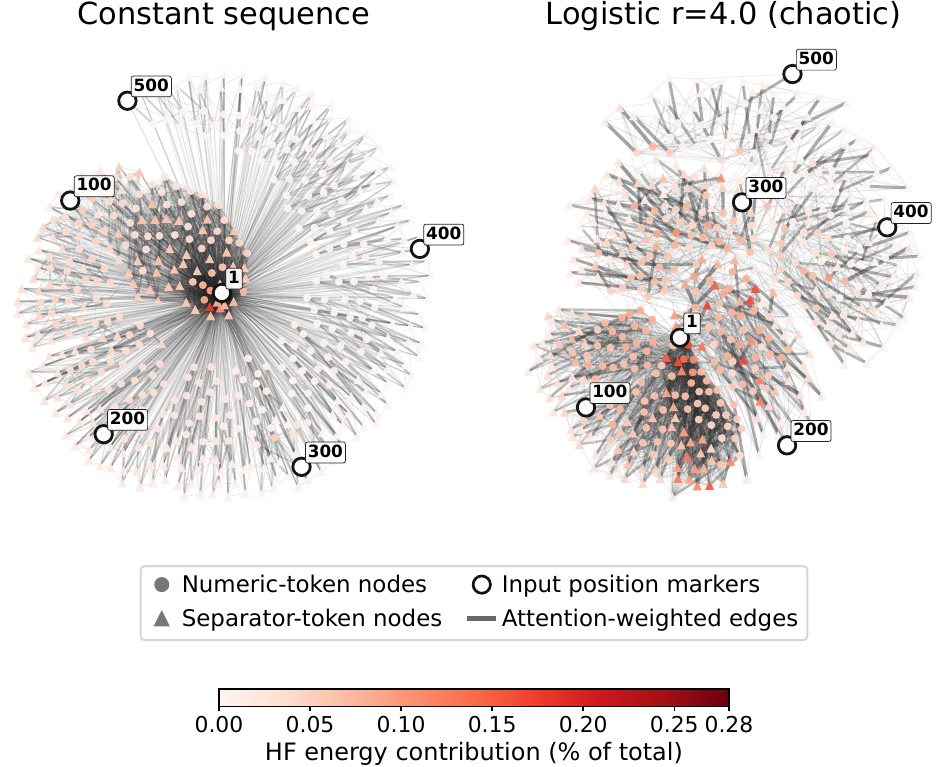}
  \caption{High-frequency hidden-state energy on attention-induced token graphs for representative constant and chaotic inputs at the final layer of Llama-3.2-3B with $N=500$. The chaotic input exhibits high-frequency energy across a broader range of token positions than the constant sequence, while both examples show stronger contributions in the early context.}
  \label{fig:high-frequency-token-graphs}
\end{figure}

\paragraph{High-Frequency Components on Token Graphs.}
Fig.~\ref{fig:high-frequency-token-graphs} shows the nodewise distribution of high-frequency hidden-state energy for representative constant and chaotic logistic-map ($r=4.0$) inputs at the final layer of Llama-3.2-3B with $N=500$. Graph construction, layout, and edge rendering follow Section~\ref{sec:graph-visualization-workflow}. Following the default HFER cutoff in Section~\ref{sec:graph-spectral-diagnostics}, we reconstruct the high-frequency signal using the 250 graph Fourier modes with the largest combinatorial-Laplacian eigenvalues. Each node is colored by the squared Euclidean norm of its reconstructed hidden-state vector, normalized by total hidden-state energy and expressed as a percentage. Both panels share a linear red color scale, with darker nodes indicating larger contributions.

Compared with the constant sequence, the chaotic logistic-map input exhibits high-frequency energy across a broader range of token positions. In both examples, contributions tend to be stronger in the early context and weaker at later positions. These observations qualitatively complement the layer-averaged HFER findings in Section~\ref{sec:spectral-diagnostics}, where additional context reduces the high-frequency energy share most strongly for simpler inputs, while chaotic inputs retain a comparatively larger share.

\begin{figure*}
  \centering\includegraphics[width=\textwidth,height=0.915\textheight,keepaspectratio]{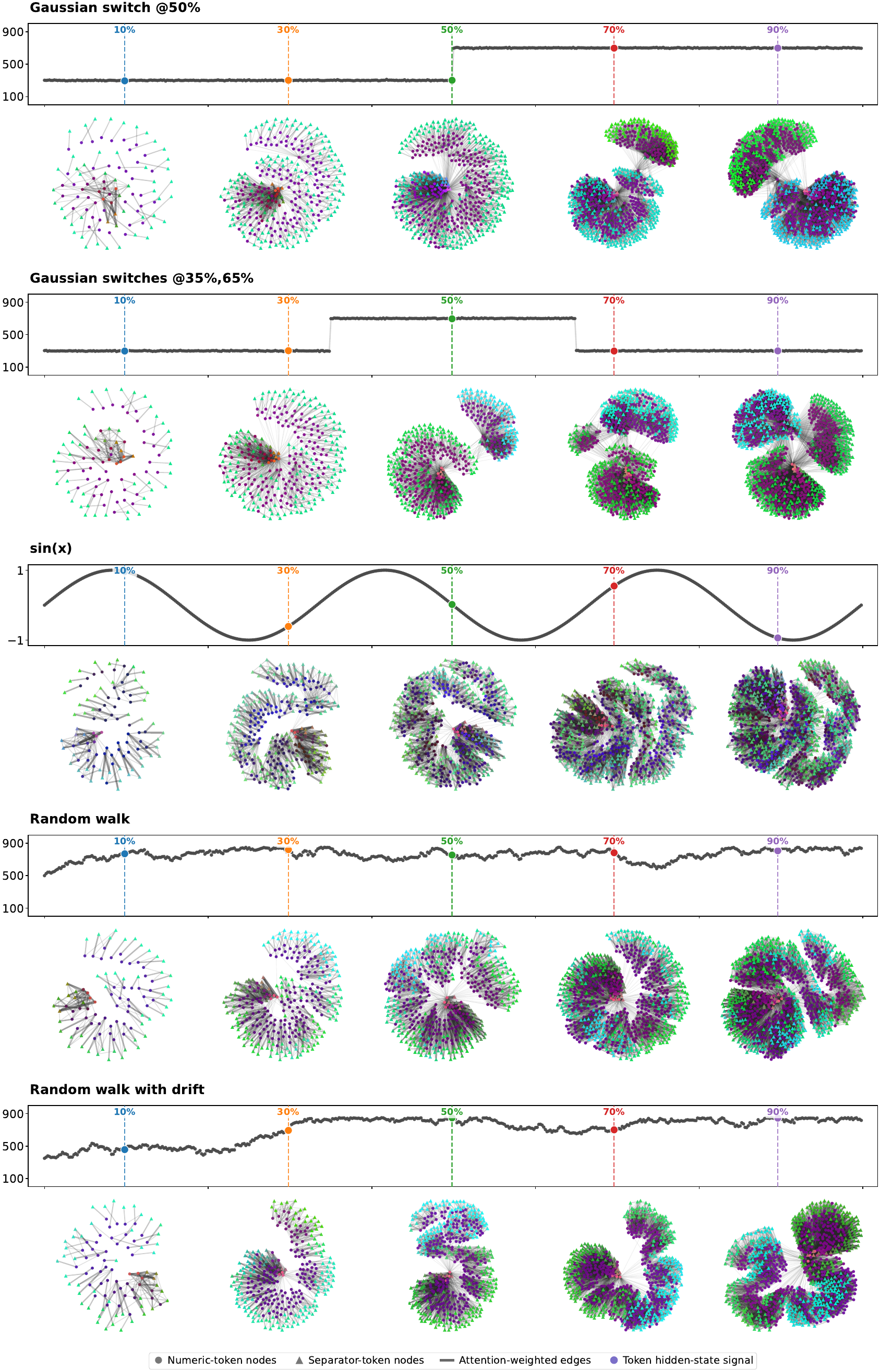}
  \caption{Last-layer attention-induced token graphs for additional structured numerical inputs with Llama-3.2-3B. Graphs are shown at 10\%, 30\%, 50\%, 70\%, and 90\% of a 1000-token reference context. All examples use the same attention-induced token-graph construction and visualization procedure as \Cref{sec:graph-visualization-workflow}. The visualizations illustrate qualitative graph-organization patterns beyond those observed for the main dynamical-system trajectories.}
  \label{fig:last-layer-additional-inputs-combined-view}
\end{figure*}

\paragraph{Additional Structured Numerical Inputs.}
Beyond the main numerical input suite introduced in \cref{sec:numerical-input-families-llm-input-preparation}, we consider several additional structured numerical sequences representing abrupt distributional shifts, smooth periodic variation, and stochastic evolution. Each full sequence contains 500 numeric tokens and 500 separator tokens, yielding a 1000-token reference context. \cref{fig:last-layer-additional-inputs-combined-view} displays the corresponding last-layer attention-induced token graphs at 10\%, 30\%, 50\%, 70\%, and 90\% of the reference context, using the same graph construction and visualization procedure described in \Cref{sec:graph-visualization-workflow}.

The Gaussian-switch inputs are piecewise Gaussian sequences with i.i.d. samples within each segment and abrupt changes in mean at prescribed context fractions. This construction adapts the distribution-switching setup used by \citet{sarfati2026shape} to study belief dynamics. In the one-switch case, samples are drawn from $\mathcal{N}(\mu=300,\sigma^2=10)$ over the first half of the sequence and from $\mathcal{N}(\mu=700,\sigma^2=10)$ over the second half. In the two-switch case, the mean is 300 over the first 35\% of the sequence, 700 over the middle 30\%, and 300 over the final 35\%, with $\sigma^2=10$ throughout. Each switch therefore has magnitude $\Delta_\mu=400$, which is large relative to the within-regime standard deviation $\sigma=\sqrt{10}$. The sampled values are rounded to the nearest integer and clipped to $[150,850]$ before serialization.

We also include a smooth periodic trajectory generated from $\sin(x)$, sampled uniformly over three full periods and rescaled to the same integer range. Unlike the periodic logistic map orbits examined in the main text, whose successive iterates can move sharply between distinct orbit values, this sequence varies smoothly from one step to the next.

Finally, we include two bounded stochastic processes: a random walk and a random walk with drift. Both follow
\[
z_{t+1}=R_{[150,850]}(z_t+\mu+\epsilon_t),
\quad
\epsilon_t\sim\mathcal{N}(0,18^2),
\]
with $\mu=0$ for the random walk and $\mu=2.5$ for the drifted random walk. The zero-drift and drifted walks are initialized at $z_0=500$ and $z_0=350$, respectively. The reflection operator $R_{[150,850]}$ maps boundary overshoots back into $[150,850]$, repeating the reflection if necessary.

\cref{fig:last-layer-additional-inputs-combined-view} suggests that attention-induced graphs reflect not only deterministic dynamical complexity, but also distributional shifts, smooth recurrence, and stochastic evolution. In the Gaussian-switch examples, the switch-defined temporal segments appear as distinct clusters in the attention-induced graphs. This observation complements \citet{sarfati2026shape}, who show that abrupt shifts in the mean of a Gaussian input distribution produce structured belief-update trajectories in hidden-state space, accompanied by corresponding changes in the model's predictive distributions over numerical outputs. Our attention-induced token-graph view provides a complementary perspective by showing that change-point structure is also reflected in attention topology as clusters aligned with the switch-defined temporal segments. The sine input produces a graph structure distinct from periodic logistic map orbits despite their shared periodicity. By contrast, the random-walk inputs produce less sharply organized graph structures, consistent with the path-dependent nature of their generating processes.

\section{Additional Controls and Higher-Dimensional Inputs}
\label[appendix]{app:additional-numerical-controls}

We complement \Cref{sec:spectral-diagnostics} by testing whether token diversity alone explains the observed organization, assessing robustness to excluding separator tokens from the analyzed graph, and examining whether the broad trends extend to higher-dimensional inputs. All experiments follow the setup in \Cref{sec:spectral-diagnostics}, except for the controls and input extensions described below.

\begin{figure*}[t]
  \centering
  \captionsetup[subfigure]{skip=2pt}
  \begin{subfigure}[t]{\textwidth}
    \centering
    \includegraphics[width=\linewidth]{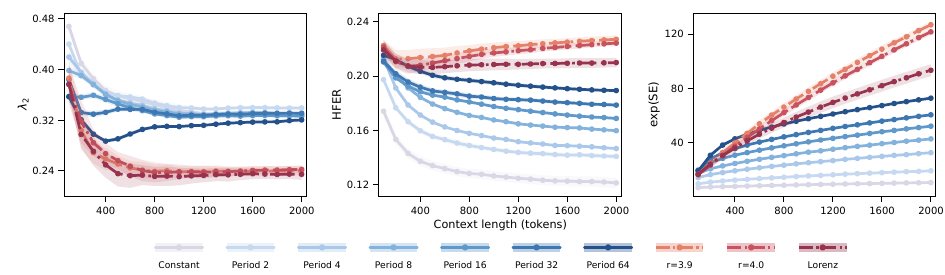}
    \caption{Main one-dimensional inputs}
    \label{fig:additional-controls-original}
  \end{subfigure}

  \begin{subfigure}[t]{\textwidth}
    \centering
    \includegraphics[width=\linewidth]{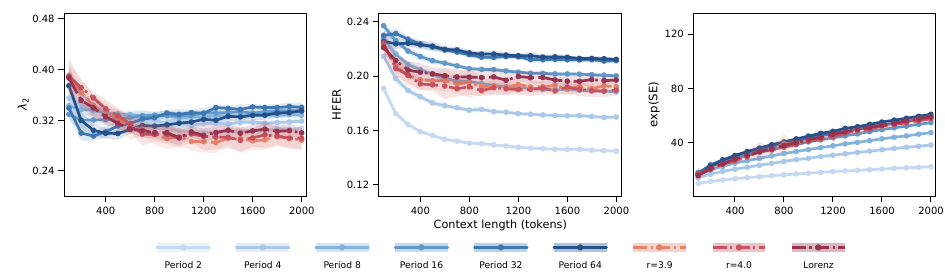}
    \caption{Shuffled numerical states}
    \label{fig:shuffled-numerical-states}
  \end{subfigure}

  \begin{subfigure}[t]{\textwidth}
    \centering
    \includegraphics[width=\linewidth]{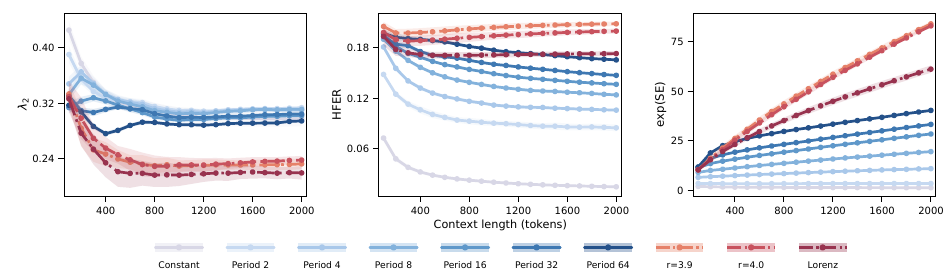}
    \caption{Analysis restricted to numerical tokens}
    \label{fig:numerical-token-subgraphs}
  \end{subfigure}

  \begin{subfigure}[t]{\textwidth}
    \centering
    \includegraphics[width=\linewidth]{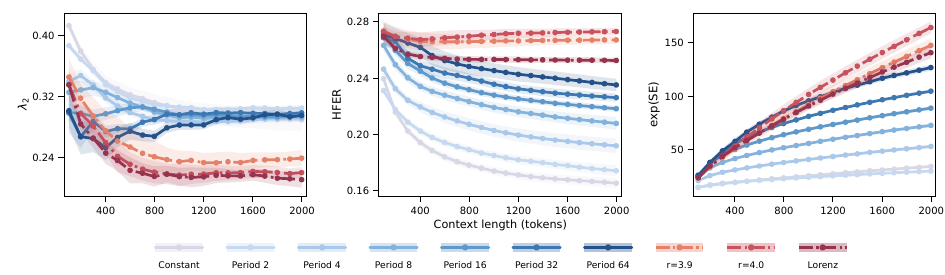}
    \caption{Three-dimensional inputs}
    \label{fig:three-dimensional-inputs}
  \end{subfigure}

  \caption{Layer-averaged graph-spectral diagnostics across context lengths under the same experimental setup as \Cref{sec:spectral-diagnostics}, except for the indicated controls and input extensions. (a) Main one-dimensional inputs provide the reference for comparison. (b) Shuffling numerical states while preserving token identities, frequencies, and separator positions substantially weakens complexity-dependent separation across all three diagnostics. (c) Restricting the analysis to numerical-token subgraphs preserves the broad qualitative trends, although with weaker separation among input families. (d) Three-dimensional constant vectors, logistic-map sequences with shared $r$ and independent initializations, and coupled Lorenz trajectories exhibit qualitatively similar context- and complexity-dependent trends to those observed in the one-dimensional setting.}
  \label{fig:additional-controls-comparison}
\end{figure*}

\paragraph{Shuffled numerical states.}
For each realization and context length, we randomly permute numerical states within the selected prefix while preserving exact token identities, frequencies, and separator positions. Constant inputs are omitted because shuffling leaves them unchanged. Although shuffling preserves each sequence's token identities and frequencies, it substantially reduces the complexity-dependent separation in all three diagnostics (\cref{fig:additional-controls-comparison}(b)). The original organization therefore cannot be accounted for by token diversity alone.

\paragraph{Numerical-token subgraphs.}
We retain the original prompts during inference, then restrict each attention-induced graph and its hidden-state signal to numerical-token positions. Selected edge weights are preserved, and degrees and Laplacians are recomputed without further attention normalization. As shown in \cref{fig:additional-controls-comparison}(c), the broad qualitative trends persist, although with weaker separation among input families, indicating that these trends are robust to excluding separator tokens from the analyzed graph.

\paragraph{Three-dimensional inputs.}
We extend the main numerical suite to constant vectors, three independent logistic-map sequences sharing the same parameter $r$, and coupled Lorenz $(x,y,z)$ trajectories. We follow the generation and quantization procedures in \cref{app:numerical-inputs-extrapolation}, with the following coordinate-wise initializations. For each realization, we independently draw $u_1,u_2,u_3\sim\mathrm{Unif}(0.3,0.7)$. These samples determine the constant-vector entries and initialize the three logistic maps, respectively, with independently randomized phases for periodic inputs. Lorenz trajectories are initialized at $(-8,7,27)+(u_1-0.5,u_2-0.5,u_3-0.5)$. Following \citet{bao2026texttrained}, we serialize three-dimensional inputs using commas to separate coordinates within each state and semicolons to mark time-step boundaries, e.g., ``\texttt{350,500,650;350,500,650;...}'' for a constant-vector sequence. As shown in \cref{fig:additional-controls-comparison}(d), three-dimensional inputs exhibit qualitatively similar context- and complexity-dependent trends to those observed in the main one-dimensional setting across all three diagnostics.

\section{Preliminary Layerwise Intervention Analysis}
\label[appendix]{app:layer-skipping}

As a first step toward connecting internal graph-spectral diagnostics to extrapolation behavior, we examine how skipping individual transformer layers affects next-state prediction error. We compare each intervention with the original model at $N=2000$ tokens, the longest context analyzed, to examine these effects in the long-context regime where the diagnostics more clearly distinguish input dynamical complexity (Section~\ref{sec:spectral-diagnostics}). For this experiment, we generate and quantize 1001 states per realization, using the first 1000 as context and the final state as the held-out next-state target.

\begin{figure*}[t]
  \centering
  \includegraphics[width=0.9\linewidth]{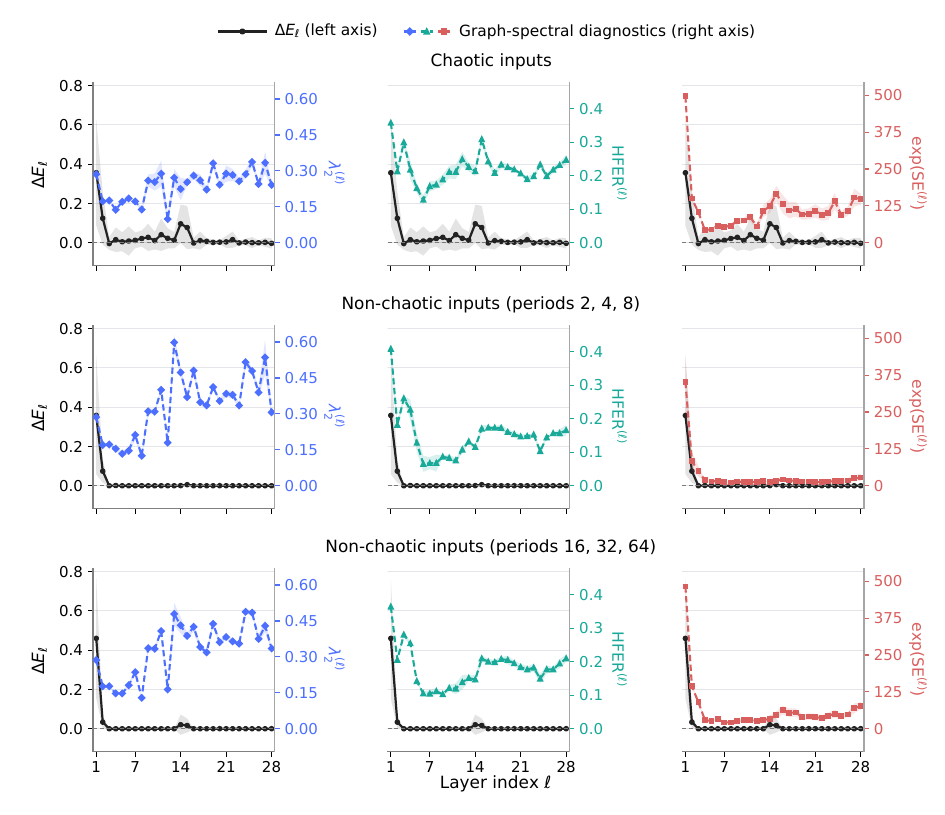}
  \caption{Preliminary layer-skipping analysis with Llama-3.2-3B at $N=2000$. Rows show chaotic inputs (logistic $r=3.9$, $r=4.0$, and Lorenz), periodic inputs with periods 2, 4, and 8, and periodic inputs with periods 16, 32, and 64, from top to bottom. Black curves show the increase in normalized absolute extrapolation error relative to the original model after skipping each layer (left axes). Colored curves show original-model normalized Fiedler values, HFER, and $\exp(\mathrm{SE})$ from left to right (right axes). Curves and shading indicate means and $\pm1$ sample standard deviation across 60 realizations pooled within each group. Extrapolation and layerwise diagnostics follow Appendices~\ref{app:numerical-inputs-extrapolation} and~\ref{app:layer-resolved-spectral-diagnostics}, respectively. Middle-layer skipping affects chaotic inputs more strongly than periodic inputs, partially aligning with spectral variation and providing initial interventional evidence linking the diagnostics to extrapolation behavior.}
  \label{fig:layer-skip-diagnostics}
\end{figure*}

For input family $f$ and realization $j$, both models receive the same context of $t$ states and are evaluated against the same held-out next-state target. Let $e^f_{j,t}$ and $e^{f,\mathrm{skip}(\ell)}_{j,t}$ denote the normalized absolute extrapolation errors of the original model and the model with layer $\ell$ bypassed, respectively. We define the paired error change as
\[
\Delta e^f_{j,t,\ell}
=
e^{f,\mathrm{skip}(\ell)}_{j,t}
-
e^f_{j,t},
\]
where positive values indicate that skipping layer $\ell$ worsens extrapolation. The prediction rule and error normalization follow Appendix~\ref{app:numerical-inputs-extrapolation}. We compare these error changes with the original model's layerwise graph-spectral diagnostics presented in Appendix~\ref{app:layer-resolved-spectral-diagnostics}. For each input group, we denote the mean of $\Delta e^f_{j,1000,\ell}$ across its 60 realizations by $\Delta E_\ell$, shown by the black curves in Figure~\ref{fig:layer-skip-diagnostics}.

Fig.~\ref{fig:layer-skip-diagnostics} shows that skipping the earliest layers produces the largest mean error increases across input groups, possibly reflecting disruption of early representations used by subsequent layers. Beyond the first two layers, error increases are most pronounced for chaotic inputs in the middle layers, particularly layers 14--15. This region broadly aligns with increasing effective spectral support, $\exp(\mathrm{SE})$, over layers 12--15 and a local HFER peak at layer 15, consistent with the broader spectral support and greater high-frequency energy observed for chaotic inputs. These observations raise the possibility that high graph-frequency components play a greater functional role in chaotic than in periodic sequence continuation, consistent with the stronger low-frequency concentration observed for periodic inputs. Later layers show smaller error increases despite elevated $\exp(\mathrm{SE})$ and continued variation in HFER, while periodic inputs exhibit weaker middle-layer effects. The alignment is therefore partial: layer skipping connects internal diagnostics to extrapolation behavior but does not isolate the causal role of spectral structure. These findings motivate future work using graph Fourier mode reconstructions and targeted frequency-band and attention-edge interventions, as discussed in Section~\ref{sec:conclusion}.

\section{Preliminary Graph-Spectral Comparison of Numerical, Natural-Language, and Code Inputs}
\label[appendix]{sec:additional-language-inputs}

\paragraph{Shared setup.}
We construct three input suites---multilingual text, source code, and English-domain text---each comprising six input types and 50 samples per type. Mirroring the numerical-input evaluation in \Cref{sec:spectral-diagnostics}, we compute the same graph-spectral diagnostics at 20 context lengths from 100 to 2000 tokens in increments of 100, using the corresponding prefixes of each 2000-token sample.

\paragraph{Multilingual text.}
The multilingual suite contains text in six languages---Arabic, Chinese, English, French, Russian, and Spanish---from MultiUN, a parallel corpus of translated United Nations documents \citep{eisele-chen-2010-multiun,tiedemann-2012-parallel}. Using English as the pivot, we align corresponding records across the six languages and concatenate consecutive aligned record sets until all six language streams contain at least 2000 tokens; we then truncate each stream to exactly 2000 tokens. The samples therefore contain aligned source material, although differences in tokenization density mean that the cutoff may fall at different positions within the final aligned record.

\paragraph{Code inputs.}
The programming-language suite comprises Python, JavaScript, Java, C++, Go, and Rust files from the StarCoder training dataset \citep{kocetkov2023stack,li2023starcoder}. For each language, we sample 50 contiguous 2000-token segments at random offsets from distinct source files. We remove encoding artifacts and invalid control characters while preserving whitespace, line breaks, indentation, comments, and other code formatting.

\paragraph{English-domain text.}
The English-domain suite comprises news from CNN/DailyMail \citep{hermann2015teaching,see-etal-2017-get}, fiction from Project Gutenberg (50 distinct books, one per sample) \citep{christou-tsoumakas-2025-artificial}, English Wikipedia articles from a November 2023 snapshot \citep{wikimedia-downloads}, scientific text from arXiv \citep{cohan-etal-2018-discourse}, dialogue from DailyDialog \citep{li-etal-2017-dailydialog}, and legal text from LexGLUE \citep{chalkidis-etal-2022-lexglue}. We apply light common preprocessing by decoding HTML entities, removing generic boilerplate-only lines, flattening internal line breaks, and collapsing repeated whitespace. When a dataset entry is shorter than 2000 tokens, we concatenate consecutive entries until reaching the target length.

\begin{figure*}
  \centering
  \begin{subfigure}[t]{\textwidth}
    \centering
    \includegraphics[width=\linewidth]{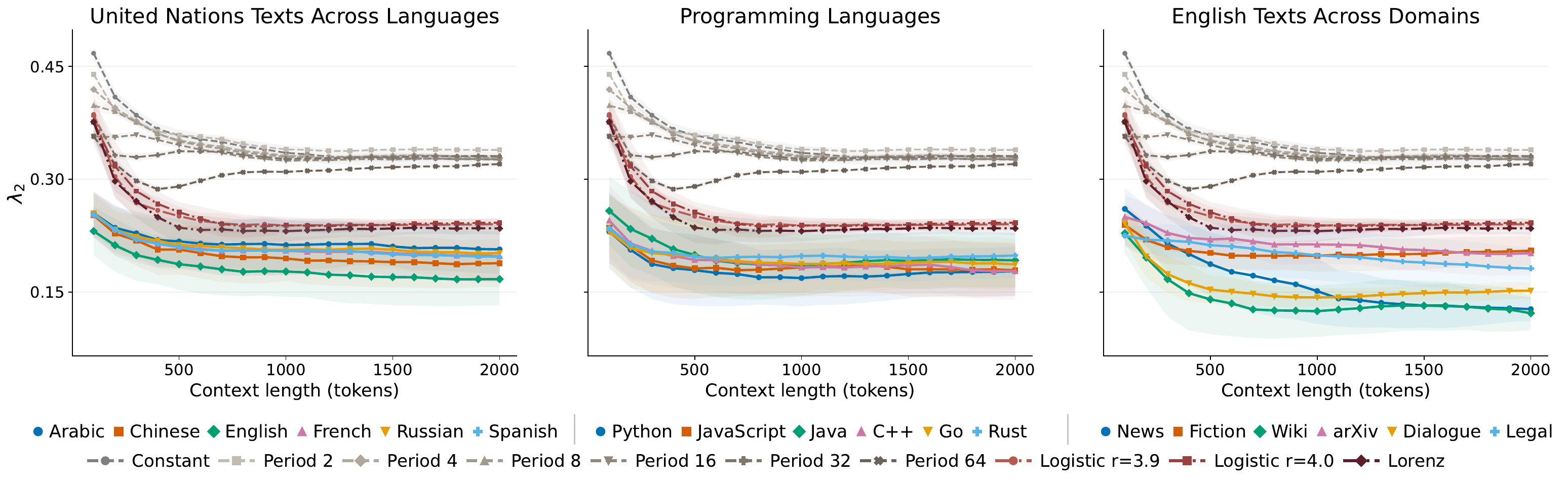}
    \caption{Normalized Fiedler value $\lambda_2$}
  \end{subfigure}

  \vspace{1em}

  \begin{subfigure}[t]{\textwidth}
    \centering
    \includegraphics[width=\linewidth]{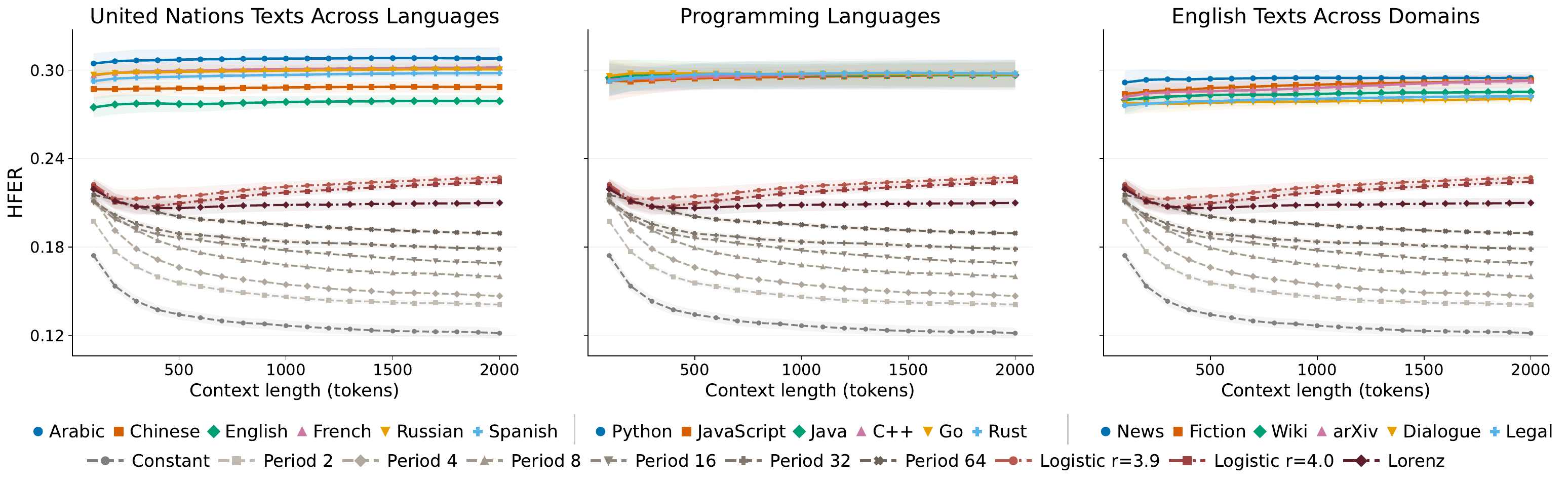}
    \caption{High-frequency energy ratio (HFER)}
  \end{subfigure}

  \vspace{1em}

  \begin{subfigure}[t]{\textwidth}
    \centering
    \includegraphics[width=\linewidth]{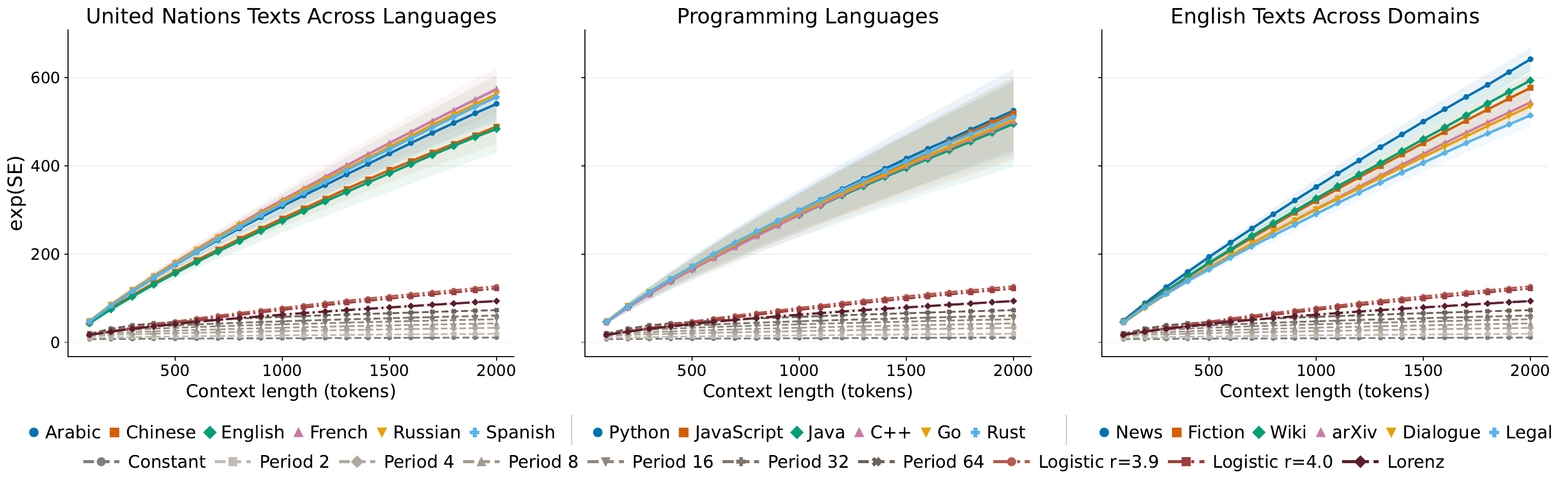}
    \caption{Effective spectral support $\exp(\mathrm{SE})$}
  \end{subfigure}
  \caption{Layer-averaged graph-spectral diagnostics for natural-language and code inputs using Llama-3.2-3B, with the main numerical families shown as dashed matched-context references. Columns compare content-aligned United Nations translations across six languages, programming-language inputs, and English-domain inputs. Curves and shading denote the mean and $\pm 1$ standard deviation across 50 samples per natural-language or code family and 20 realizations per numerical family. Across context lengths, the natural-language and code family means exhibit lower normalized Fiedler values, higher HFER, and broader effective spectral support than the numerical inputs; this coarse separation corresponds to weaker global attention connectivity, a larger high-frequency share of hidden-state energy, and a broader distribution of that energy across graph Fourier modes.}
  \label{fig:language-combined-diagnostics}
\end{figure*}

\paragraph{Findings and open questions.}
\cref{fig:language-combined-diagnostics} reveals a consistent separation of the natural-language and code inputs from the controlled numerical inputs across all three diagnostics. Relative to the numerical inputs, the natural-language and code inputs exhibit lower normalized Fiedler values, indicating less globally integrated and more localized attention topology, together with higher HFER, indicating sharper hidden-state variation over that topology. The separation is especially pronounced in effective spectral support, with $\exp(\mathrm{SE})$ remaining substantially larger for the natural-language and code families than for the numerical families across all evaluated context lengths. This difference indicates that their hidden-state energy is distributed across a substantially larger effective number of graph Fourier modes. These results support a coarse representational distinction between controlled numerical sequences and natural-language or code inputs. More controlled comparisons are needed to determine which input and model properties contribute to the coarse separation of natural-language and code inputs from numerical inputs and whether meaningful graph-spectral orderings exist within the multilingual-text, source-code, and English-domain suites.

\end{document}